\documentclass[10pt,journal,onecolumn]{IEEEtran}
\usepackage{amsmath,cite,psfrag,subfigure,graphicx,url,color}
\usepackage{algorithm}
\usepackage{multirow}
\usepackage{multicol}
\usepackage{algorithmic}

\newtheorem{theorem}{{\textbf{Theorem}}}[section]

\newtheorem{property}[theorem]{\textbf{Property}}

\newenvironment{proof}[1][Proof]{\begin{trivlist}
\item[\hskip \labelsep {\bfseries #1}]}{\end{trivlist}}
\newenvironment{definition}[1][Definition]{\begin{trivlist}
\item[\hskip \labelsep {\bfseries #1}]}{\end{trivlist}}

\newcommand{\qed}{\nobreak \ifvmode \relax \else
      \ifdim\lastskip<1.5em \hskip-\lastskip
      \hskip1.5em plus0em minus0.5em \fi \nobreak
      \vrule height0.75em width0.5em depth0.25em\fi}

\begin{document}
\graphicspath{{figures/}}
\newcommand{\sps}{\scriptsize}
\newcommand{\eqs}{\scriptsize}
\newcommand{\Tabsize}{\scriptsize}
\title{Decorrelation of Neutral Vector Variables:\\
Theory and Applications}
\author{Zhanyu Ma,~\IEEEmembership{Member,~IEEE}, Jing-Hao Xue,~Arne Leijon, Zheng-Hua Tan,~\IEEEmembership{Senior Member,~IEEE},\\
 Zhen Yang,~\IEEEmembership{Member,~IEEE},~and Jun Guo
%
%
%
%
\thanks{Z. Ma and J. Guo are with Pattern Recognition and Intelligent System Lab., Beijing University of Posts and Telecommunications, Beijing, China.}
\thanks{J.-H. Xue is with the Department of Statistical Science, University College London, London, United Kingdom.}
\thanks{A. Leijon is with the School of Electrical Engineering, KTH - Royal Institute of Technology, Stockholm, Sweden.}
\thanks{Z.-H. Tan is with the Department of Electronic Systems, Aalborg University, Aalborg, Denmark.}
\thanks{Z. Yang is with the College of Computer Science, Beijing University of Technology, Beijing, China.}
\thanks{The corresponding author is Z. Ma. Email: mazhanyu@bupt.edu.cn}
}

\maketitle

\begin{abstract}
In this paper, we propose novel strategies for neutral vector variable decorrelation. Two
fundamental invertible transformations, namely serial nonlinear transformation and parallel nonlinear transformation,
are proposed to carry out the decorrelation. For a neutral vector variable, which is not multivariate Gaussian distributed, the conventional principal component analysis (PCA) cannot yield mutually independent scalar variables. With the two proposed transformations, a highly negatively correlated neutral vector can be transformed to a set of mutually independent scalar variables with the same degrees of freedom. We also evaluate the decorrelation performances for the vectors generated from a single Dirichlet distribution and a mixture of Dirichlet distributions. The mutual independence is verified with the distance correlation measurement. The advantages of the proposed decorrelation strategies are intensively studied and demonstrated with synthesized data and practical application evaluations.
\end{abstract}


\IEEEdisplaynotcompsoctitleabstractindextext

\IEEEpeerreviewmaketitle
\section{Introduction}
In many pattern recognition and machine learning areas, Gaussian distributions, among other probability distributions, have been ubiquitously applied to describe data distribution, with the assumption that these data are Gaussian distributed~\cite{Park2013}. However, in many applications the distribution of data is asymmetric or constrained~\cite{Nguyen2013}. For example, the pixel values in a color or grey image~\cite{Ma2011a,Bouguila2006}, the ratings assigned to an item in collaborative filtering~\cite{Ma2015,Salakhutdinov2008,Salakhutdinov2008a}, and the epigenetic mark values in epigenome-wide-association studies~\cite{Ji2005,Ma2013a} have strictly bounded support (\emph{e.g.}, $x\in[0,c]$). In speech enhancement, the spectrum coefficients~\cite{Mohammadiha2013,Mohammadiha2013a} are semi-bounded (\emph{i.e.}, $x\in(0,+\infty$)). The $l_2$ norms of the spatial fading correlation~\cite{Mammasis2009} and the yeast gene expressions~\cite{Taghia2014} are equal to $1$ and such data convey directional property (\emph{i.e.}, $\|\mathbf{x}\|_2=1$). A common property of the aforementioned data is that, these data have~\emph{not only} a specific support range,~\emph{but also} a non-bell distribution shape. Apparently, these properties do not match the natural properties of a Gaussian distribution (\emph{i.e.}, the definition domain is unbounded and the distribution shape is symmetric). Therefore, such data are non-Gaussian distributed~\cite{Ma2011}. It has been demonstrated in many recent studies that explicitly utilizing the non-Gaussian characteristics can significantly improve the performance in practice~\cite{Ma2011,Ma2011a,Bouguila2006,Ji2005,Ma2013a,Jung2014,Mohammadiha2013,Mammasis2009,Mohammadiha2013a,Taghia2014,Zao2012,Xu2014,Si2012,Si2013,Chen2016}.

One typical type of non-Gaussian distributed data, among others, is the one that represents proportions. In the
frequently used mixture modeling technique~\cite{Bishop2006,McLachlan2000,Ma2011a}, the weighting factors denote the proportions of each mixture
component in the whole mixture model. In the text mining area, the Dirichlet distribution is used to model topic relations,~\emph{i.e.}, the proportions with which a specific topic appears in the total set of documents~\cite{Blei2004,Blei2006,Blei2012}. For analyzing color images, the normalized RGB space, which is often used as pure color space by discarding the illuminance~\cite{Wyszecki2000,Bouguila2007,Boutemedjet2009,Rana2015}, represents the proportions of RGB channels in the whole color space. In time series signal processing~\cite{Subramaniam2003,Yedlapalli2010}, the difference between two adjacent line spectral frequencies (LSFs) conveys
the proportion of frequency distance (in angle) to half of the unit circle's circumference. The LSFs are less sensitive to quantization noise than other representations and are widely used in speech coding~\cite{Subramaniam2003,Ma2013,Ma2014}. Also, the parameters in the multinomial
distribution~\cite{Forbes2011,Chen2015} represent probabilities for each particular event to happen in the trial sequence. In~\cite{Chen2013}, a novel online kernel learning algorithm,
called QKRLS, was developed, which is computationally efficient and can be used for online regression and classification.

Data representing proportions can be denoted by a $K+1$ dimensional vector $\mathbf{x}=[x_1,\ldots,x_K,$ $x_{K+1}]^{\text{T}}$
with $K$ degrees of freedom. Each element $x_k$ is nonnegative and the sum of all the elements in $\mathbf{x}$ is a
constant (usually can be normalized to $1$). Connor et al.~\cite{Connor1969} introduced the concept ``neutrality'' to
investigate a particular type of independence for the elements in $\mathbf{x}$. Even though the resulting neutral vector represents a
particular type of independence after a substraction-normalization operation~\cite{James1972}, the elements in the neutral vector are mutually
highly correlated, or rather, negatively correlated. Intuitively, if one proportion increases, then the remaining proportions
would decrease correspondingly, since the summation of all the proportions is a constant.

For correlated random vector variables, principal component analysis (PCA) is a popular technique used for applications such as
data decorrelation, dimension reduction, lossy data compression, and feature extraction~\cite{Bishop2006,Torre2012,Han2014}. It is also known
as Karhunen-Lo\`{e}ve  transform (KLT) in transform coding~\cite{Chen2012,Torun2013}. It can be considered as an orthogonal transformation of the correlated variables into a set of uncorrelated scalar variables, which are named as principal components. This transformation is linear and invertible. PCA is the optimal decorrelation strategy for multivariate-Gaussian distributed data~\cite{Bishop2006}. For data from a multivariate-Gaussian distribution, the resulting transformed scalar variables are not only mutually uncorrelated but also mutually independent. For data from other sources, PCA can only guarantee that the scalar variables are mutually uncorrelated.

{Independent component analysis (ICA) is a computational method applied to separate a multivariate vector variable into a set of additive and mutually independent scalar variables (sources)~\cite{Stone2004,Nguyen2011}. With the assumption that the source signals are independent of each other and the source signals are non-Gaussian distributed, ICA attempts to decorrelate a multivariate vector variable into mutually independent non-Gaussian scalar variables. ICA can be applied to several fields such as face recognition~\cite{Kwak2007}, blind source separation~\cite{Santamari2013}, and wireless communications~\cite{Arnaut2014}.}

A neutral vector has a bounded support (in $[0,c]$) and is negatively correlated (the off-diagonal elements in the covariance matrix is negative). Thus, it cannot follow a multivariate-Gaussian distribution. In this case, applying PCA to neutral vector can only yield mutually uncorrelated but not mutually independent scalar variables. {Although ICA can yield mutually independent non-Gaussian scalar variables, it cannot preserve the bounded support property.} By considering the neutrality, the highly correlated variables in a neutral vector can be decorrelated into a set of independent variables with nonlinear transformation. Moreover, such procedure does not depend on the eigenvalue decomposition of the covariance matrix.

In this paper, we propose two fundamental transformation strategies, namely the serial nonlinear transformation (SNT) and the parallel nonlinear transformation (PNT),
to decorrelate neutral vectors. These invertible nonlinear transformations take the advantages of the completely neutrality. We prove that the above mentioned nonlinear
transformations can decorrelate the neutral vector variable into a set of mutually independent variables. Particularly, if the
neutral vector variable is Dirichlet distributed, each of the transformed variables follows the beta distribution, which is
actually a special case of the Dirichlet distribution with two parameters.

{Although nonlinear kernel functions can be introduced to carry out kernel PCA~\cite{Scholkopf1997,Varon2015} or kernel ICA~\cite{Bach2002,Xiao2015} such that the vector variable decorrelation can be implemented in a nonlinear manner, the proposed nonlinear transformation strategies are different from these ones. In kernel PCA, input vectors are firstly mapped into a feature space via a kernel function, and then the standard PCA is applied to conduct the decorrelation~\cite[Ch.~12.3]{Bishop2006}. Similar approaches are applied to kernel ICA. Therefore, kernel PCA and kernel ICA each contain two stages, which are nonlinear kernel mapping and linear decorrelation (in the feature space). In contrast to this, the proposed nonlinear transformation strategies (\emph{i.e.}, SNT and PNT) do not require kernel mapping. It is a one-stage nonlinear operation in the decorrelation implementation.}

For a neutral random vector, the decorrelation strategies are based on each observed vector~\emph{only} and does not require any statistical
information (\emph{e.g.}, the covariance matrix) of the whole observation set. In other words, the decorrelation strategies are model independent. Therefore, the proposed decorrelation strategies reduce the computational complexity, compared with PCA which requires eigenvalue decomposition of the covariance matrix. {ICA has even higher computational costs than PCA.} The decorrelation of a vector variable is important and very helpful in many applications (\emph{e.g.}, source coding, dimension reduction, and feature selection~\cite{Wang2016,Yuan2016}). Hence, the proposed
decorrelation strategies are novel and useful for the data with neutrality.

The rest of this paper is organized as follows: we review the concept of neutrality in Sec.~\ref{Sec: Neutral Vector Variable}. The proposed transformation strategies are introduced in Sec.~\ref{Sec:NLT} where the proof of mutually independence is also provided. In Sec.~\ref{Sec:Dirichlet}, we take the Dirichlet distribution as an example for neutral vectors. Comprehensive evaluations of the proposed strategies with synthesized and real data are presented in Sec.~\ref{Sec:Exp and Disc}. We draw some conclusions in Sec.~\ref{Sec:Conclusion}.

\section{Neutral Vector Variable}
\label{Sec: Neutral Vector Variable}
Assuming we have a random vector variable $\mathbf{x}=[x_1,x_2,$ $\ldots,x_K,x_{K+1}]^{\text{T}}$, where $x_k> 0$ and $\sum_{k=1}^{K+1}x_k=1$.
Let $\mathbf{x}_{k1}=[x_1,\ldots,x_k]^{\text{T}}$ and $\mathbf{x}_{k2}=[x_{k+1},\ldots,x_{K+1}]^{\text{T}}$. The vector $\mathbf{x}_{k1}$ is neutral if $\mathbf{x}_{k1}$ is independent of $\mathbf{w}_k=\frac{1}{1 - s_k}\mathbf{x}_{k2}$, for $1\leq k\leq K$~\cite{Connor1969,James1980}, where $s_k=\sum_{i=1}^k x_i$ and $s_0=0$. If for all $k$, $\mathbf{x}_{k1}$ are neutral, then $\mathbf{x}$ is defined as a \emph{completely neutral} vector~\cite{Connor1969,Hankin2010}. A neutral vector with $(K+1)$ elements has $K$ degrees of freedom.

The idea of neutrality was introduced by Connor et al.~\cite{Connor1969} for describing constrained variables with the property
mentioned above. It was originally developed for biological applications. According to the above definition, the
neutral vector conveys a particular type of independence among its elements, even though the element variables themselves are
mutually negatively correlated. A complete neutral vector variable has a set of properties, we list those will be used in this paper here:
\begin{property}[Mutually Independence]
\label{Mutually Independence}
For completely neutral vector $\mathbf{x}$, define $z_k=\frac{x_k}{1-s_{k-1}}$ and $z_1=x_1$, we have $z_1, z_2,\ldots,z_K$ are mutually independent.
\end{property}

\begin{property}[Aggregation Property]
\label{Aggregation Property}
For a completely neutral vector $\mathbf{x}$, when adding any adjacent elements $x_r$ and $x_{r+1}$ together, the resulting $K$-dimensional vector $\mathbf{x}^{r\uplus r+1}=[x_1,\ldots,x_r+x_{r+1},\ldots,x_{K+1}]$ is a completely neutral vector again.
\end{property}
\begin{proof}
Due to the completely neutral property, we have $\mathbf{x}_{k1}\perp \mathbf{w}_{k}$, $1\leq k\leq K$, where $\perp$ denotes independence. For the $K$-dimensional vector $\mathbf{x}^{r\uplus r+1}$\footnotemark\footnotetext{We use similar notation as defined at the beginning of Sec.~\ref{Sec: Neutral Vector Variable}.},
\begin{enumerate}
\item When $1\leq k<r$, it can be recognized that the elements in $\mathbf{x}^{r\uplus r+1}_{k1}$ are identical to those in $\mathbf{x}_{k1}$. The only difference between $\mathbf{w}^{r\uplus r+1}_{k}$ and $\mathbf{w}_{k}$ is that $\mathbf{w}^{r\uplus r+1}_{k}$ contains element $\frac{x_r+x_{r+1}}{1-s_{k}}$ while $\mathbf{w}_{k}$ contains $[\frac{x_r}{1-s_{k}}, \frac{x_{r+1}}{1-s_{k}}]$. Based on these facts, we can immediately show that $\mathbf{x}^{r\uplus r+1}_{k1}$ is independent of all the elements in $\mathbf{w}^{r\uplus r+1}_{k}$ except for $\frac{x_r+x_{r+1}}{1-s_{k}}$. On the other hand, we also have
    \begin{eqnarray}
    \eqs
    \mathbf{x}^{r\uplus r+1}_{k1}\perp \mathbf{w}_{k}
    \Rightarrow & \mathbf{x}^{r\uplus r+1}_{k1}\perp [\frac{x_{r}}{1-s_{k}},\frac{x_{r+1}}{1-s_{k}}]\\
    \Rightarrow & \mathbf{x}^{r\uplus r+1}_{k1}\perp \frac{x_{r}+x_{r+1}}{1-s_{k}},
    \end{eqnarray}
    Hence, it can be proved that $\mathbf{x}^{r\uplus r+1}_{k1}$ is independent of $\frac{x_r+x_{r+1}}{1-s_{k}}$ and, therefore, $\mathbf{x}^{r\uplus r+1}_{k1}$ is neutral for $1\leq k <r$.

\item When $r< k< K$, $\mathbf{w}^{r\uplus r+1}_{k}=\mathbf{w}_{k}$ and the distinct elements in $\mathbf{x}^{r\uplus r+1}_{k1}$ and $\mathbf{x}_{k1}$ are $x_r+x_{r+1}$ and $[x_r, x_{r+1}]$, respectively. For the same reasoning, we can also prove that $\mathbf{x}^{r\uplus r+1}_{k1}$ is neutral for $r< k\leq K$.

%
\end{enumerate}
Based on these, we conclude that $\mathbf{x}^{r\uplus r+1}_{k1}$ is neutral for $1\leq k\leq K$ and $\mathbf{x}^{r\uplus r+1}$ is completely neutral.
\end{proof}

Usually, the dimensions in a completely neutral vector should be equally treated. In other words, the positions of the dimensions do not affect the properties of the vector. In order to explicitly convey this fact, we make the following definition:
\begin{definition}
For a~\emph{completely neutral} vector $\mathbf{x}$, if~\emph{arbitrarily} permuted version of $\mathbf{x}$ is still~\emph{completely neutral}, then this vector is~\emph{exchangeably completely neutral}.
\end{definition}

In the field of statistical analysis, a typical
variable which has the above mentioned properties is the Dirichlet variable. For the Bayesian analysis of mixture
models~\cite{Bishop2006,McLachlan2000}, the weighting factors of the mixture components are usually modelled by a Dirichlet distribution.
Recently, the Dirichlet process (\emph{e.g.},~\cite{Blei2005,Orbanz2010}) was applied for nonparametric Bayesian analysis. If we
represent the Dirichlet process  with the so-called stick-breaking process~\cite{Blei2005}, the independence among different
generating steps can be expressed explicitly as a neutral vector with infinite dimensionality. The Dirichlet process is the
cornerstone of non-parametric Bayesian analysis and applied to a variety of practical signal  and feature analysis problems. Thus, the concept
of neutral vectors is very useful in many signal processing, pattern recognition, and other practical applications.

\section{Transformations for Neutral Vectors}
\label{Sec:NLT}

\begin{algorithm}[!t]
   \caption{Serial Nonlinear Transformation}
   \label{alg:SNT}
\Tabsize
\begin{algorithmic}
   \STATE {\bfseries Input:} Neutral vector $\mathbf{x}=[x_1,\ldots,x_K,x_{K+1}]^{\text{T}}$
   \STATE Set $\mathbf{x}_1 = \mathbf{x}$, $i=1$
     \REPEAT
    \STATE Assign the value of the $1$st element of $\mathbf{x}_i$ to $u_i$;
    \STATE $i=i+1$, $\mathbf{x}_i=\mathbf{x}_{i-1}$, with the first element in $\mathbf{x}_{i-1}$ removed;
    \STATE Normalize the remaining elements in $\mathbf{x}_i$ as $\mathbf{x}_i = \mathbf{x}_i /\| \mathbf{x}_i\|_1$
   \UNTIL{$i==K$}
   \STATE {\bfseries Output:} Transformed vector $\mathbf{u}= [u_1,\ldots,u_K]^{\text{T}}$.

\end{algorithmic}
   \end{algorithm}
In most signal processing applications, the transformations we use are linear or linear according to some
nonlinear kernel functions. Even though we could apply PCA directly to the neutral random vector variable, this linear
transformation could only decorrelate the data, but cannot guarantee the independence if the data are not Gaussian. Furthermore, PCA does not exploit the neutrality~\cite{Ma2011}. In this case, PCA is not optimal for decorrelating neutral vectors. By considering the exchangeably complete neutrality, we propose two nonlinear invertible transformations, namely the serial nonlinear transformation (SNT) and the parallel nonlinear transformation (PNT). Each of the proposed nonlinear transformations can decorrelate the vector variable into a set of mutually independent variables. In contrast to PCA, the transformations do not require any statistical information (\emph{e.g.}, the covariance matrix) of the observed vector set. Thus, it avoids the eigenvalue analysis for PCA and, therefore, the computational complexity is reduced.

\subsection{Serial Nonlinear Transformation}

\begin{figure}[!t]
\vspace{-0mm}

\psfrag{N}[c][c]{ \tiny$\mathcal{N}$}

\vspace{0mm}
          \centering
          \includegraphics[width=.3\textwidth]{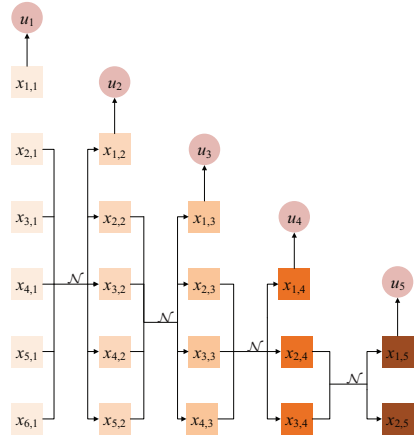}
          \vspace{-2mm}
          \caption{\label{SNT}\footnotesize An example of SNT with $K=5$. The transformed variables
          are $u_1 ={x_{1,1}}$, $u_2 =x_{1,2}$, $u_3=x_{1,3}$, $u_4=x_{1,4}$ and
          $u_5=x_{1,5}$. $x_{i,j}$ denotes the $i^{th}$ element in $\mathbf{x}_j$. $\mathcal{N}$ is the $l_1$-norm normalization.}
          \vspace{-5mm}
\end{figure}
For an exchangeable completely neutral vector variable with $K$ degrees of freedom, if we process the vector variable with the SNT strategy described in
Algorithm~\ref{alg:SNT}, the $K+1$ dimensional vector variable $\mathbf{x}$ is transformed to a vector $\mathbf{u}$ with $K$
variables. These $K$ scalar variables are mutually independent.

The proof of mutually independence of $\mathbf{u}$ is straightforward. In the transformed vector $\mathbf{u}$, the first element $u_1=x_1=z_1$. The second element of $\mathbf{u}$ is
\begin{equation}
\eqs
u_2 =\frac{x_2}{1-s_1}=z_2.
\end{equation}
Similarly, we have
\begin{equation}
\eqs
u_3 = \frac{\frac{x_3}{1-s_1}}{1-\frac{x_2}{1-s_1}}=\frac{x_3}{1-s_2}=z_3.
\end{equation}
More generally, we can obtain that
\begin{equation}
\label{Eq:XXX}
\eqs
u_k = \frac{\frac{x_k}{1-s_{k-2}}}{1-\frac{x_{k-1}}{1-s_{k-2}}}=\frac{x_k}{1-s_{k-1}}=z_k,\ \ 3\leq k\leq K.
\end{equation}
According to Property~\ref{Mutually Independence}, Eq.~\ref{Eq:XXX} shows that the $K$ variables in $\mathbf{u}$ are mutually independent. Since the SNT is invertible and $\mathbf{x}$ has $K$ degrees of freedom, $\mathbf{u}$ contains the same amount
of information as $\mathbf{x}$. An example of the SNT with $K=5$ is illustrated in Fig.~\ref{SNT}.

   \vspace{-0mm}
\subsection{Parallel Nonlinear Transformation}
\begin{figure}[!t]
\vspace{-0mm}
\psfrag{R}[c][c]{ \tiny$\mathcal{R}$}

\vspace{0mm}
          \centering
          \includegraphics[width=.3\textwidth]{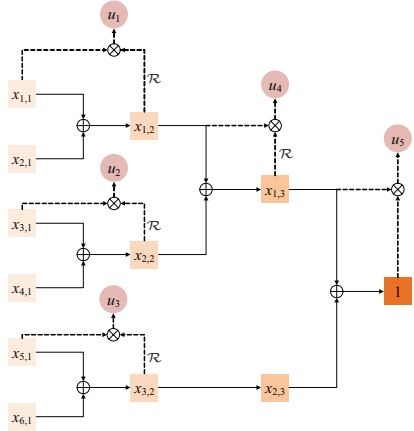}
          \vspace{-2mm}
          \caption{\label{PNT}\footnotesize An example of PNT with $K=5$. The transformed coefficients
          are $u_1 ={x_{1,1}}/x_{1,2} $, $u_2 =x_{3,1}/x_{2,2}$, $u_3=x_{5,1}/x_{3,2}$, $u_4=x_{1,2}/x_{1,3}$ and
          $u_5=x_{1,3}$. $\mathcal{R}$ represents the reciprocal operation.}
          \vspace{-5mm}
\end{figure}

The SNT algorithm needs $K$ rounds of iterations to finalize the transformation. In order to facilitate the operation, we can also carry out the nonlinear transformation in a
parallel way, which is referred as the parallel nonlinear transformation (PNT). The PNT scheme is introduced in Alg.~\ref{alg:PNT}. At each
iteration, the dimension of the processed vector is reduced by half. Finally, we still get a vector variable
$\mathbf{u}$ with $K$ mutually independent element variables. An example of PNT with $K=5$ is shown in Fig.~\ref{PNT}. The proofs
of independence are as follows.
\subsubsection{Independence within Subvector $\mathbf{u}_i$}
\label{Sec:Subvector Mutual Independence}
According to the PNT scheme in Alg.~\ref{alg:PNT}, at the $i^{th}$ iteration, we obtain a new vector $\mathbf{x}_i = [x_{1,i-1}+x_{2,i-1}, x_{3,i-1}+x_{4,i-1}, \ldots]^{\text{T}}$, where we denote the $l^{th}$ element in the $\mathbf{x}_i$ as $x_{l,i}$ and define $\mathbf{x}_1 = \mathbf{x}$. With Property~\ref{Aggregation Property} (the aggregation property), it can be readily shown that $\mathbf{x}_i$ is completely neutral for any $i$.

In the $i^{th}$ iteration, the elements in $\mathbf{u}_i$ are $u_{l,i}=\frac{x_{2l-1,i}}{x_{2l-1,i}+x_{2l,i}}$. For any two elements $u_{m,i}$ and $u_{n,i}$ (we assume $m<n$ here), we have the following relation
\begin{equation}
\eqs
\label{Eq:Mutual Independence}
[\ldots,x_{2m-1,i},x_{2m,i}]^{\text{T}} \perp [\ldots, w_{2n-1,i},w_{2n,i},\ldots]^{\text{T}},
\end{equation}
which is due to the completely neutrality of $\mathbf{x}_{i}$. Here, $w_{2n-1,i}=\frac{x_{2n-1,i}}{1-s_{2m}}$. By recognizing $u_{m,i} = \frac{x_{2m-1,i}}{x_{2m-1,i}+x_{2m,i}}$ and $u_{n,i} = \frac{x_{2n-1,i}}{x_{2n-1,i}+x_{2n,i}}=\frac{w_{2n-1,i}}{w_{2n-1,i}+w_{2n,i}}$ and denoting $\bar{u}_{m,i}=1-u_{m,i}$ and $\bar{u}_{n,i}=1-u_{n,i}$, the relation between $[u_{m,i},\bar{u}_{m,i},u_{n,i},\bar{u}_{n,i}]^{\text{T}}$ and $[x_{2m-1,i},x_{2m,i},w_{2n-1,i},w_{2n,i}]^{\text{T}}$ can be presented as
\begin{equation}
\eqs
\begin{split}
&[u_{m,i},\bar{u}_{m,i},u_{n,i},\bar{u}_{n,i}]^{\text{T}}= \mathcal{H}\left([x_{2m-1,i},x_{2m,i},w_{2n-1,i},w_{2n,i}]^{\text{T}}\right).
\end{split}
\end{equation}

\begin{algorithm}[!t]
   \caption{Parallel Nonlinear Transformation~\cite{Ma2013}}
   \label{alg:PNT}
\Tabsize
\begin{algorithmic}
                 \STATE {\bfseries Step 1}. Initialization
                 \STATE Set $\mathbf{x}_1=\mathbf{x}$, $i=2$
                 \STATE {\bfseries Step 2}. Aggregation
                 \STATE $L = \textnormal{length}(\mathbf{x}_{i-1})-1$
                 \IF {$L$ is even}
                 \FOR  {$l=1,l\leq L/2,l++$}
                 \STATE $x_{l,i} = x_{2l-1,i-1}+x_{2l,i-1}$
                 \STATE $u_{l,i-1} = \frac{x_{2l-1,i-1}}{x_{l,i}}$
                 \ENDFOR
                 \STATE $\mathbf{x}_i = [x_{1,i},\ldots,x_{l,i},x_{L+1,i-1}]^T$
                 \STATE $\mathbf{u}_{i-1} = [u_{1,i-1},\ldots,u_{l,i-1}]^T$
                 \ELSE
                 \FOR {$l=1,l<(L+1)/2,l++$}
                 \STATE $x_{l,i} = x_{2l-1,i-1}+x_{2l,i-1}$
                 \STATE $u_{l,i-1} = \frac{x_{2l-1,i-1}}{x_{l,i}}$
                 \ENDFOR
                 \STATE $\mathbf{x}_i = [x_{1,i},\ldots,x_{l,i}]^T$\\
                 \STATE $\mathbf{u}_{i-1} = [u_{1,i-1},\ldots,u_{l,i-1}]^T$\\
                 \ENDIF
                 \STATE {\bfseries Step 3}. Stop criterion
                 \IF {$\textnormal{length}(\mathbf{x}_{i})==2$}
                 \STATE $\mathbf{u}_i={x}_{1,i}$, go to step $4$
                 \ELSE
                 \STATE $i=i+1$, go to step $2$.
                 \ENDIF
                 \STATE {\bfseries Step 4}. Return the transformed coefficients $\mathbf{u} = [\mathbf{u}_1^T,\ldots,\mathbf{u}_{i}^T]^T$.\\
\end{algorithmic}
   \end{algorithm}
The Jacobian matrix of the above transformation is
\begin{equation}
\eqs
\begin{split}
\mathcal{J}_{\mathcal{H}}=\left[\begin{array}{cc}\mathbf{A} & \mathbf{0}\\\mathbf{0} & \mathbf{B}\end{array}\right],
\end{split}
\end{equation}
where
\begin{equation}
\eqs
\begin{split}
\mathbf{A}=\left[\begin{array}{cc}\frac{\partial u_{m,i}}{\partial x_{2m-1,i}}&\frac{\partial u_{m,i}}{\partial x_{2m,i}}\\\frac{\partial \bar{u}_{m,i}}{\partial x_{2m-1,i}} & \frac{\partial \bar{u}_{m,i}}{\partial x_{2m,i}}\end{array}\right]\ \text{and}\ \mathbf{B}=\left[\begin{array}{cc}\frac{\partial u_{n,i}}{\partial w_{2n-1,i}}&\frac{\partial u_{n,i}}{\partial w_{2n,i}}\\\frac{\partial \bar{u}_{n,i}}{\partial w_{2n-1,i}} & \frac{\partial \bar{u}_{n,i}}{\partial w_{2n,i}}\end{array}\right].
\end{split}
\end{equation}
By the principles of variable substitution, we have
\begin{equation}
\eqs
\label{Eq:Joint-1}
\begin{split}
&f(x_{2m-1,i},x_{2m,i},w_{2n-1,i},w_{2n,i})\\
=&\mid \det (\mathcal{J}_{\mathcal{H}}) \mid f(u_{m,i},\bar{u}_{m,i},u_{n,i},\bar{u}_{n,i})\\
=&\mid \det (\mathbf{A})\mid\mid \det (\mathbf{B})\mid f(u_{m,i},\bar{u}_{m,i},u_{n,i},\bar{u}_{n,i}).
\end{split}
\end{equation}
Similarly, the following relations also hold
\begin{equation}
\eqs
\label{Eq:Joint-2}
\begin{split}
f(x_{2m-1,i},x_{2m,i})=&\mid \det (\mathbf{A})\mid f(u_{m,i},\bar{u}_{m,i})\\
f(w_{2n-1,i},w_{2n,i})=&\mid \det (\mathbf{B})\mid f(u_{n,i},\bar{u}_{n,i}).
\end{split}
\end{equation}
Combining~\eqref{Eq:Mutual Independence},~\eqref{Eq:Joint-1}, and~\eqref{Eq:Joint-2}, we can obtain
\begin{equation}
\eqs
\begin{split}
f(u_{m,i},\bar{u}_{m,i},u_{n,i},\bar{u}_{n,i}) = f(u_{m,i},\bar{u}_{m,i})f(u_{n,i},\bar{u}_{n,i})
\end{split}
\end{equation}
and infer that $u_{m,i}\perp u_{n,i}$.
Hence, the elements within the group $\mathbf{u}_i$ are mutually independent. Note that this proof is different from that shown in~\cite{Ma2013}, as no permutation property of $\mathbf{x}$ is used.

\subsubsection{Independence between Subvectors $\mathbf{u}_i$ and $\mathbf{u}_j$}
In Algorithm~\ref{alg:PNT}, each iteration yields one subvector $\mathbf{u}_i$ based on $\mathbf{x}_i$. Taking
two arbitrary  subvectors $\mathbf{u}_i$ and $\mathbf{u}_j$ (we suppose $i<j$) and selecting arbitrary elements $u_{p,i}$ and $u_{q,j}$ from
each subvector, respectively, we have the following transformation
\begin{equation}
\eqs
\begin{split}
[u_{p,i},u_{q,j},\bar{u}_{q,j}]^{\text{T}} = \mathcal{G}\left([u_{p,i},x_{2q-1,j},x_{2q,j}]^{\text{T}}\right),
\end{split}
\end{equation}
where
$u_{q,j} = \frac{x_{2q-1,j}}{x_{2q-1,j}+x_{2q,j}}$ and $\bar{u}_{q,j} = 1 - u_{q,j}$.
Similar as the proof procedure in Sec.~\ref{Sec:Subvector Mutual Independence}, we get the Jacobian matrix of the transformation $\mathcal{G}$ as
\begin{equation}
\eqs
\mathcal{J}_{\mathcal{G}} = \left[\begin{array}{ccc}1 & 0 & 0\\0 & \multicolumn{2}{c}{\multirow{2}{*}{{$\mathbf{C}$}}}\\ 0 & & \end{array}\right],\ \mathbf{C}=\left[\begin{array}{cc}\frac{\partial u_{q,j}}{\partial x_{2q-1,j}}&\frac{\partial u_{q,j}}{\partial x_{2q,j}}\\\frac{\partial \bar{u}_{q,j}}{\partial x_{2q-1,j}} & \frac{\partial \bar{u}_{q,j}}{\partial x_{2q,j}}\end{array}\right].
\end{equation}

With the fact\footnotemark\footnotetext{This can be directly observed from the exchangeably complete neutrality.} that $u_{p,i}\perp[x_{2q-1,j},x_{2q,j}]^{\text{T}}$, we have
\begin{equation}
\eqs
\label{XXX}
\begin{split}
f(u_{p,i},u_{q,j},\bar{u}_{q,j}) =& \frac{1}{ \mid \det(\mathbf{C})\mid}f(u_{p,i},x_{2q-1,j},x_{2q,j})\\
 =& \frac{1}{ \mid \det(\mathbf{C})\mid} f(u_{p,i})f(x_{2q-1,j},x_{2q,j}).
 \end{split}
\end{equation}
In addition to this, we also have
\begin{equation}
\eqs
\label{YYY}
f(x_{2q-1,j},x_{2q,j}) = \mid \det(\mathbf{C})\mid f(u_{q,j},\bar{u}_{q,j}).
\end{equation}
Thus, substituting~\eqref{YYY} into~\eqref{XXX}, we finally get
\begin{equation}
\eqs
f(u_{p,i},u_{q,j},\bar{u}_{q,j})=f(u_{p,i})f(u_{q,j},\bar{u}_{q,j}),
\end{equation}
which indicates $u_{p,i}\perp u_{q,j}$. Then it can be concluded that any two subvectors are mutually independent.

Combining the conclusion of independence within and among the subvectors,
the mutual independence of the element variables in $\mathbf{u}$ is proved.

\subsubsection{Fast Parallel Nonlinear Transformation}
According to Alg.~\ref{alg:PNT}, the implementation of PNT needs to check if the length of $\mathbf{x}_i$ in each iteration is even or odd. This is due to the fact that the number of elements in $\mathbf{x}$ is not always equal to power of $2$.  Inspired by the fast Fourier transform~\cite{Brigham2002}, we design a fast PNT (FPNT) algorithm to facilitate the practical computation with zero-padding. Zero-padding is a technique usually employed to make the length of a vector equal to a power of $2$, by adding zeros to the end of the vector so that the total number of elements equals the next higher power of $2$. The vector $\mathbf{x}$ is expanded with zero-padding to the next higher power of $2$. During each iteration in the transformation, the vector length reduces to half, until the length of the vector reduces to two. This algorithm skips the check of parity, and, therefore, the practical computational time is reduced. It is convenient to implement in practice. It is worthy to note that this FPNT algorithm has similar computational complexity to the PNT flow chart shown in Alg.~\ref{alg:PNT}. The FPNT algorithm is introduced in Algorithm~\ref{alg:FPNT}.
\begin{algorithm}[!t]

   \caption{Fast Parallel Nonlinear Transformation}
   \label{alg:FPNT}
\Tabsize
\begin{algorithmic}
   \STATE {\bfseries Input:} Neutral vector $\mathbf{x}=[x_1,\ldots,x_K,x_{K+1}]^{\text{T}}$
   \STATE Set $T=\lceil\log_2 \left(K+1\right)\rceil$ and $P=2^{T}-(K+1)$
   \STATE Set $\mathbf{x}_{\text{zp}}=[\mathbf{x}^{\text{T}},\mathbf{0}_{P}^{\text{T}}]^{\text{T}}$\ (zero-padding) $^{\dag}$
   \STATE Set $\mathbf{x}_1 = \mathbf{x}_{\text{zp}}$
    \FOR {$t=1,t\leq T,t++$}
     \STATE $\mathbf{u}_{t}^{\text{temp}} = {\mathbf{x}_t^{\text{odd}}}./({\mathbf{x}_t^{\text{odd}}+\mathbf{x}_t^{\text{even}}})$ $^{\ddag}$
     \STATE Set $\mathbf{u}_{t}$ to be a vector containing only the elements that are not equal to one in $\mathbf{u}_{t}^{\text{temp}}$
     \STATE $\mathbf{x}_{t+1}=\mathbf{x}_t^{\text{odd}}+\mathbf{x}_t^{\text{even}}$
    \ENDFOR
   \STATE {\bfseries Output:} Transformed vector $\mathbf{u} = [\mathbf{u}_1^{\text{T}},\ldots,\mathbf{u}_{T}^{\text{T}}]^{\text{T}}$.
\end{algorithmic}
$^{\dag}$ $\mathbf{0}_{P}$ is a $P\times 1$ vector contains only $0$.\\
$^{\ddag}$ $\mathbf{x}_t^{\text{odd}}$ and $\mathbf{x}_t^{\text{even}}$ represent the odd and even elements in $\mathbf{x}_t$, respectively. The operator $./$ denotes element-wise division. Moreover, we define $\frac{0}{0}=1$.
\end{algorithm}

\section{Dirichlet Variable: An Example}
\label{Sec:Dirichlet}
In the above nonlinear transformations, we did not assign any explicit distribution to the neutral vector variable. Indeed, the
transformation itself does not require us to know the specific distribution of the vector variable, with the assumption that the vector variable is exchangeably completely neutral. In this section, we will take the
Dirichlet variable as an intuitive example. It has been showed in~\cite{James1980} that the Dirichlet distribution is characterized by neutrality and a vector drawn from a Dirichlet distribution is~\emph{completely} neutral. Moreover, any permutation of such vector (which is generated from a Dirichelt distribution) is also a~\emph{completely} neutral vector (\emph{i.e.}, exchangeably completely neutral). Note that, a~\emph{completely} neutral vector~\emph{may not} have such permutation property~\cite{Connor1969}.

The Dirichlet density function is defined as
\begin{equation}
\eqs
\mathbf{Dir}(\mathbf{x};\boldsymbol \alpha)\! =\! \frac{\Gamma(\sum_{k=1}^{K+1}\alpha_k)}{\prod_{k=1}^{K+1}\Gamma(\alpha_k)}  \prod_{k=1}^{K+1}x_{k}^{\alpha_k-1},x_k\!\geq\!0, \sum_{k=1}^{K+1}x_k\!=\!1,\alpha_k\!>\!0.
\end{equation}
If we take any element $x_k$ from $\mathbf{x}$ and denote the remaining normalized elements as $\mathbf{x}_{\setminus
k}=\frac{1}{1-x_k}[x_1,\ldots,$ $x_{k-1},x_{k+1},\ldots,$ $x_{K+1}]^{\text{T}}$, it can be shown that~\cite{Frigyik2010}
\begin{equation}
\eqs
f(x_k,\mathbf{x}_{\setminus k})\!=\!\mathbf{Beta}(x_k;\alpha_k,\!\sum_{i=1,i\neq k}^{K+1}\alpha_i)\times\mathbf{Dir}(\mathbf{x}_{\setminus k};\boldsymbol \alpha_{\setminus k}),
\end{equation}
where $\boldsymbol \alpha_{\setminus k}=[\alpha_1,\ldots,\alpha_{k-1},\alpha_{k+1},\ldots,\alpha_{K+1}]^{\text{T}}$ and
\begin{equation}
\eqs
\mathbf{Beta}(x;a,b) = \frac{\Gamma(a+b)}{\Gamma(a)\Gamma(b)}x^{a-1}(1-x)^{b-1}
\end{equation}
is the beta density function, which is exactly a Dirichlet density function with two parameters $a$ and
$b$. Thus a Dirichlet variable $\mathbf{x}=[x_1,\ldots,x_K,x_{K+1}]^{\text{T}}$ is a neutral vector. Furthermore, the Dirichlet variable
has the aggregation property as~\cite{Frigyik2010}
\begin{equation}
\eqs
\mathbf{x}_{i+j}\sim\mathbf{Dir}(\mathbf{x}_{i+j};\boldsymbol \alpha_{i+j}),
\end{equation}
where $\mathbf{x}_{i+j}=[x_1,\ldots,x_i + x_j,\ldots,x_{K+1}]^{\text{T}}$ and $\boldsymbol \alpha_{i+j}=[\alpha_1,\ldots,\alpha_i +
\alpha_j,\ldots,\alpha_{K+1}]^{\text{T}}$. These properties can be easily shown by the principles of variable substitution.

For the SNT strategy, the transformed variable $u_k$ is beta distributed as
\begin{equation}
\eqs
u_k \sim \mathbf{Beta}(u_k;\alpha_k, \sum_{i=k+1}^{K+1}\alpha_i),
\end{equation}
which can be proved by the neutrality and the aggregation properties.

For each loop in the PNT algorithm (Algorithm~\ref{alg:PNT}), we define a new parameter vector $\boldsymbol \alpha_i$ for the
$i$th loop ($i\geq 2$). The update rule for $\boldsymbol \alpha_i$ is the same as $\mathbf{x}_i$ and $\boldsymbol \alpha_1 =
\boldsymbol \alpha$. In the $i$th loop, we can obtain a Dirichlet distribution by aggregating the elements
$x_{3,i-1},\ldots,x_{L+1,i-1}$ together as
\begin{equation}
\eqs
\begin{split}
&[x_{1,i-1},x_{2,i-1},\sum_{l=3}^{L+1}x_{l,i-1}]^{\text{T}}\\
\sim & \mathbf{Dir}(x_{1,i-1},x_{2,i-1},\sum_{l=3}^{L+1}x_{l,i-1};\alpha_{1,i-1},\alpha_{2,i-1},\sum_{l=3}^{L+1}\alpha_{l,i-1}).
\end{split}
\end{equation}
By considering that $\sum_{l=3}^{L+1}x_{l,i-1}$ is a neutral variable, the normalized version of the remaining two variables
$x_{1,i-1},x_{2,i-1}$ are again Dirichlet distributed with two parameters. This is equivalent to a beta distribution. Thus the
obtained coefficient $u_{1,i-1}=x_{1,i-1}/(x_{1,i-1}+x_{2,i-1})$ follows a beta distribution as
\begin{equation}
\label{PDF4Beta}
\eqs
u_{1,i-1} \sim \mathbf{Beta}(u_{1,i-1};\alpha_{1,i-1}, \alpha_{2,i-1}).
\end{equation}
Based on the same reasoning, we can show that $u_{l,i-1}$ is also beta distributed. Thus, with SNT or PNT, the Dirichlet variable can
be decorrelated into a vector with the same degrees of freedom. Due to the complete neutrality, the element variables in the transformed vector are mutually
independent, and each element variable is beta distributed.

\section{Experimental Results and Discussions}
\label{Sec:Exp and Disc}
\vspace{0mm}
The importance of independence arises in many applications. The proposed nonlinear transformation methods can decorrelate a neutral vector variable into a set of mutually independent scalar variables. In order to illustrate the decorrelation performance, the distance correlation (DC)~\cite{Szekely2007,Szekely2009}, which measures statistical dependence between two random variables, is applied to evaluate the mutual independence of the scalar variables after transformation. Unlike the commonly used Pearson correlation coefficient~\cite{Pearson1895,Wilcox2005}, the DC is zero if and only if the random variables are statistically mutually independent~\cite{Szekely2012}. Given a set of paired samples $(X_n,Y_n),\ n=1,\ldots,N$, all pairwise Euclidean distances $a_{ij}$ and $b_{ij}$ are calculated as
\begin{equation}
\eqs
a_{ij} = \|X_i-X_j\|,\ \ \ b_{ij} = \|Y_i-Y_j\|,\ \ \ i,j = 1,\ldots, N.
\end{equation}
Taking the doubly centered distances, we have
\begin{equation}
\eqs
\begin{split}
A_{ij} &= a_{ij}-\bar{a}_{i\cdot}-\bar{a}_{\cdot j}+\bar{a}_{\cdot\cdot},\ B_{ij} = b_{ij}-\bar{b}_{i\cdot}-\bar{b}_{\cdot j}+\bar{b}_{\cdot\cdot},
\end{split}
\end{equation}
where $\bar{a}_{i\cdot}$ denotes the mean of the $i$th row, $\bar{a}_{\cdot j}$ is the mean of the $j$th column, and $\bar{a}_{\cdot\cdot}$ stands for the grand mean of the matrix. The same definitions apply to $\bar{b}_{i\cdot}$, $\bar{b}_{\cdot j}$, and $\bar{b}_{\cdot\cdot}$.

In order to evaluate the statistical significance of the DC, a permutation test is employed. The $p$-value for the permutation test is calculated as follows:
\begin{enumerate}
\item For the original data $(X_n, Y_n)$, create a new data set $(X_n, Y_{n^*})$, where $n^*$ denotes a permutation of the set $\{1,\ldots,N\}$. The permutation set is selected randomly as drawing without replacement;
\item Calculate a DC for the randomized data£»
\item Repeat the above two steps a large number of times, the $p$-value for this permutation test is the proportion of the DC values in step $2$ that are larger than the DC from the original data.
\end{enumerate}
The null hypothesis in this case is that the two variables involved are independent of each other (the DC is $0$). When the corresponding $p$-value is smaller than $0.05$, the null-hypothesis is rejected so that these two variables are~\emph{not} independent (but could still be uncorrelated). Hence, $p$-value greater than $0.05$ indicates independence.
We choose the significance level as $0.05$ in the remaining parts of this paper.

In this section, we firstly compare PNT/SNT with PCA and ICA, with evaluation of decortication performance. Next, we demonstrate the decorrelation performance of PNT (in terms of mutual independence) with both synthesized and real data. Afterwards, we apply the proposed strategy to real-life applications to improve corresponding practical performance.

\subsection{Comparisons of SNT, PNT, PCA, and ICA}
\subsubsection{Computational Complexity}
\label{Sec:Computational Complexity}
In practical applications, the computational complexity of decorrelation is usually a concern. We now analyze the computational complexities of SNT and PNT, respectively, and compare them with that of the conventionally used PCA and ICA strategies.

\begin{itemize}
\item  {SNT and PNT}\\
As described in Algorithm~\ref{alg:SNT}, each iteration yields one element in the target vector $\mathbf{u}$. Hence, when decorrelating a ($K+1$) neutral vector variable (with $K$ degrees of freedom) into a set of $K$ independent scalar variables, $K$ iterations are required. During each iteration, one summation and $L$ division should be operated for the purpose of normalization, where $L$ is the number of elements in the intermediate vector $\mathbf{x}_i$. Therefore, if we treat the summation as one floating-point operation and the
division as eight times of that\footnotemark\footnotetext{According to T. Minka's Lightspeed Matlab toolbox~\cite{Minka}\ \url{http://research.microsoft.com/en-us/um/people/minka/software/lightspeed/}.}, the computational complexity for SNT is $\mathcal{O}(NK^2)$.

When applying Algorithm~\ref{alg:FPNT} to decorrelate the neutral vector in a parallel manner, at most $\lceil\log_2 \left(K+1\right)\rceil$ iterations are required. Within each iteration, about $L/2$ summations and $L/2$ divisions with an even $L$ or $(L+1)/2$
summations and $(L+1)/2$ divisions with an odd $L$ are needed. Therefore, with the same consideration of the floating-point operation above, the computational complexity for PNT is $\mathcal{O}(NK\log K)$, since $L=K$ at the first iteration and $L$ will reduce to (approximately) half in each of the consequent iteration.

With the above analysis, we can conclude that the PNT algorithm is more efficient than the SNT algorithm and preferable in practice, although both algorithms can nonlinearly transform the neutral vector into a set of mutually independent scalars.

\item {PCA}\\
The operation of PCA includes two parts: 1) eigenvalue analysis of the covariance matrix and 2) decorrelation of the vector. Many approaches exist for an eigenvalue analysis. To our best knowledge, the fastest method so-far is the method proposed by Luk et al.~\cite{Luk2003}. The computational cost is about $\mathcal{O}(K^2\log K)$ for a $K\times K$ covariance matrix. For the decorrelation, multiplying the source vector with the eigenvector matrix will have computational cost around $\mathcal{O}(K^2)$. Therefore, the computational cost for PCA is, on average, $\mathcal{O}(NK^2\log K)$.

Hence, the proposed SNT- and PNT-based decorrelation methods are more efficient than the PCA-based method.
  \item {ICA}\\
Although robust source separation performance can be achieved by ICA, the drawback of algorithms for carrying out ICA is the high computational complexity~\cite{Shwartz2004}. Typical algorithms for ICA requires centering, whitening, and dimension reduction as preprocessing steps to facilitate the calculation. Unlike PNT/SNT or PCA which converges fast, the convergence of ICA also depends on the number of iterations. Hence, analytically tractable solution does not exist. As introduced in~\cite{Laparra2011}, the computational cost for ICA, with $M$ iterations, is $\mathcal{O}(MNK^2)$
\end{itemize}

\begin{table*}[!t]
\vspace{-5mm}
\caption{\label{Tab:SingleDecorrelation}\footnotesize Evaluation of the decorrelation performance on the data generated from a Dirichlet distribution with $\boldsymbol \alpha = [2,5,6,3,7]^{\text{T}}$. The null hypothesis is that the related two dimensions are independent from each other (\emph{i.e.}, the DC is $0$). The first row: $p$-values for the generated data. The second row: $p$-values for the decorrelated data via PNT. The third row: $p$-values for the decorrelated data via PCA. The fourth row: $p$-values for the decorrelated data via ICA. The $p$-values that are smaller than $0.05$ are marked with underline, indicating that the corresponding two random variables are not independent.}
\centering
\subtable[\label{Tab:ORG1}\sps$N=100$, original.]
{\Tabsize \begin{tabular} {c|@{}c@{}c@{}c@{}c@{}}

                & \ \ $x_1$\ \  & \ \ $x_2$\ \  & \ \ $x_3$ \ \ & \ \ $x_4$\ \ \\
   \hline
          $x_1$ & \ \ $0$ \ \      & \ \ $0.198$\ \  & \ \ $0.127$\ \     & \ \ $0.376$\ \ \\
          $x_2$ &       &\ \ $0$   \ \    &\ \ $\underline{0.007}$ \ \   & \ \ $0.140$\ \ \\
          $x_3$ &       &       & \ \ $0$   \ \     & \ \ ${0.067}$\ \ \\
          $x_4$ &       &       &       &\ \  $0$  \ \
   \end{tabular}}
   \qquad
   \hspace{-7mm}
\subtable[\label{Tab:ORG2}\sps$N=200$, original.]
 {\Tabsize \begin{tabular} {c|@{}c@{}c@{}c@{}c@{}}

                &\ \  $x_1$\ \  &\ \  $x_2$\ \  &\ \  $x_3$\ \  &\ \  $x_4$\ \ \\
   \hline
          $x_1$ & \ \  $0$   \ \    & \ \ ${0.054}$\ \   & \ \ ${0.063}$ \ \  &\ \  $0.189$\ \ \\
          $x_2$ &       &\ \  $0$    \ \  &\ \ $\underline{0.001}$ \ \   &\ \  $\underline{0.024}$\ \ \\
          $x_3$ &       &       &\ \  $0$   \ \     &\ \  $\underline{0.047}$\ \ \\
          $x_4$ &       &       &       & \ \ $0$  \ \
   \end{tabular}}
   \qquad   \hspace{-7mm}
\subtable[\label{Tab:ORG3}\sps$N=400$, original.]
{\Tabsize \begin{tabular} {c|@{}c@{}c@{}c@{}c@{}}

                &\ \  $x_1$\ \  &\ \  $x_2$ \ \ &\ \  $x_3$\ \  &\ \  $x_4$\ \ \\
   \hline
          $x_1$ & \ \ $0$    \ \    &\ \  $\underline{0.010}$ \ \  &\ \  $\underline{0.004}$ \ \  &\ \  ${0.069}$\ \ \\
          $x_2$ &       &\ \ $0$   \ \    &\ \ $\underline{0.000}$\ \    &\ \  $\underline{0.001}$\ \ \\
          $x_3$ &       &       &\ \ $0$   \ \      &\ \  $\underline{0.002}$\ \ \\
          $x_4$ &       &       &       &\ \ $0$   \ \
   \end{tabular}}
   \qquad   \hspace{-7mm}
   \subtable[\label{Tab:ORG4}\sps$N=800$, original.]
{\Tabsize \begin{tabular} {c|@{}c@{}c@{}c@{}c@{}}

                & \ \ $x_1$\ \  &\ \  $x_2$\ \  &\ \  $x_3$\ \  &\ \  $x_4$\ \ \\
   \hline
          $x_1$ & \ \ $0$     \ \   & \ \ $\underline{0.000}$ \ \  & \ \ $\underline{0.00}0$\ \   & \ \ $\underline{0.007}$\ \ \\
          $x_2$ &       &\ \   $0$ \ \    &\ \ $\underline{0.000}$\ \    & \ \ $\underline{0.000}$\ \ \\
          $x_3$ &       &       &\ \ $0$   \ \     & \ \ $\underline{0.000}$\ \ \\
          $x_4$ &       &       &       & \ \ $0$   \ \
   \end{tabular}}
       \qquad\hspace{-7mm}
   \subtable[\label{Tab:PNT1}\sps$N=100$, with PNT.]
{\Tabsize \begin{tabular} {c|@{}c@{}c@{}c@{}c@{}}

                &\ \  $u_1$\ \  & \ \ $u_2$\ \  &\ \  $u_3$\ \  & \ \ $u_4$\ \ \\
   \hline
          $u_1$ & \ \  $0$    \ \   &\ \  $0.455$\ \   &\ \  $0.426$\ \   &\ \  $0.546$\ \ \\
          $u_2$ &       &\ \ $0$   \ \    &\ \ $0.481$ \ \   &\ \  $0.405$\ \ \\
          $u_3$ &       &       &\ \   $0$  \ \     &\ \  $0.495$\ \ \\
          $u_4$ &       &       &       & \ \ $0$  \ \
   \end{tabular}}
   \qquad\hspace{-7mm}
   \subtable[\label{Tab:PNT2}\sps$N=200$, with PNT.]
{\Tabsize \begin{tabular} {c|@{}c@{}c@{}c@{}c@{}}

                & \ \ $u_1$ \ \ & \ \ $u_2$ \ \ &\ \  $u_3$ \ \ &\ \  $u_4$\ \ \\
   \hline
          $u_1$ &  \ \ $0$    \ \   &\ \  $0.464$\ \   &\ \  $0.527$\ \   &\ \  $0.455$\ \ \\
          $u_2$ &       &\ \ $0$  \ \     &\ \ $0.621$ \ \   & \ \ $0.625$\ \ \\
          $u_3$ &       &       &\ \ $0$    \ \     & \ \ $0.508$\ \ \\
          $u_4$ &       &       &       &\ \ $0$   \ \
   \end{tabular}}
   \qquad\hspace{-7mm}
\subtable[\label{Tab:PNT3}\sps$N=400$, with PNT.]
{\Tabsize \begin{tabular} {c|@{}c@{}c@{}c@{}c@{}}

                &\ \  $u_1$\ \  &\ \  $u_2$\ \  &\ \  $u_3$\ \  &\ \  $u_4$\ \ \\
   \hline
          $u_1$ & \ \ $0$    \ \    &\ \  $0.583$\ \   &\ \  $0.484$\ \   & \ \ $0.668$\ \ \\
          $u_2$ &       &\ \ $0$    \ \    &\ \ $0.538$\ \    &\ \  $0.402$\ \ \\
          $u_3$ &       &       & \ \ $0$   \ \     &\ \  $0.582$\ \ \\
          $u_4$ &       &       &       & \ \ $0$   \ \
   \end{tabular}}
   \qquad\hspace{-7mm}
\subtable[\label{Tab:PNT4}\sps$N=800$, with PNT.]
{\Tabsize \begin{tabular} {c|@{}c@{}c@{}c@{}c@{}}

                & \ \ $u_1$\ \  & \ \ $u_2$\ \  &\ \  $u_3$ \ \ & \ \ $u_4$\ \ \\
   \hline
          $u_1$ & \ \ $0$    \ \    &\ \  $0.519$\ \   &\ \  $0.360$ \ \  &\ \  $0.367$\ \ \\
          $u_2$ &       &\ \ $0$   \ \    &\ \ $0.561$\ \    &\ \  $0.496$\ \ \\
          $u_3$ &       &       &\ \ $0$     \ \    &\ \  $0.564$\ \ \\
          $u_4$ &       &       &       &\ \  $0$  \ \
   \end{tabular}}
       \qquad\hspace{-7mm}
   \subtable[\label{Tab:PCA1}\sps$N=100$, with PCA.]
{\Tabsize \begin{tabular} {c|@{}c@{}c@{}c@{}c@{}}

                & \ \ $u_1$\ \  & \ \ $u_2$\ \  & \ \ $u_3$\ \  & \ \ $u_4$\ \ \\
   \hline
          $u_1$ & \ \   $0$  \ \    & \ \ $0.307$ \ \  & \ \ $0.565$\ \   &\ \  $0.606$\ \ \\
          $u_2$ &       &\ \  $0$   \ \   &\ \ $0.211$ \ \   &\ \  $0.330$\ \ \\
          $u_3$ &       &       &\ \  $0$    \ \    & \ \ $0.207$\ \ \\
          $u_4$ &       &       &       & \ \  $0$ \ \
   \end{tabular}}
   \qquad\hspace{-7mm}
   \subtable[\label{Tab:PCA2}\sps$N=200$, with PCA.]
{\Tabsize \begin{tabular} {c|@{}c@{}c@{}c@{}c@{}}

                & \ \ $u_1$\ \  &\ \  $u_2$ \ \ & \ \ $u_3$\ \  & \ \ $u_4$\ \ \\
   \hline
          $u_1$ &  \ \ $0$     \ \   &\ \  $0.142$ \ \  &\ \  $0.511$ \ \  & \ \ $0.625$\ \ \\
          $u_2$ &       &\ \  $0$  \ \    &\ \ ${0.075}$\ \    & \ \ $0.152$\ \ \\
          $u_3$ &       &       & \ \ $0$     \ \    &\ \  $\underline{0.019}$\ \ \\
          $u_4$ &       &       &       &\ \ $0$   \ \
   \end{tabular}}
   \qquad\hspace{-7mm}
\subtable[\label{Tab:PCA3}\sps$N=400$, with PCA.]
{\Tabsize \begin{tabular} {c|@{}c@{}c@{}c@{}c@{}}

                & \ \ $u_1$ \ \ & \ \ $u_2$\ \  &\ \  $u_3$\ \  &\ \  $u_4$\ \ \\
   \hline
          $u_1$ &  \ \   $0$   \ \ & \ \ $\underline{0.048}$ \ \  &\ \  $0.395$\ \   &\ \  $0.472$\ \ \\
          $u_2$ &       &\ \ $0$   \ \    &\ \ $\underline{0.003}$ \ \   &\ \  ${0.084}$\ \ \\
          $u_3$ &       &       & \ \ $0$   \ \     & \ \ $\underline{0.000}$\ \ \\
          $u_4$ &       &       &       & \ \ $0$   \ \
   \end{tabular}}
   \qquad\hspace{-7mm}
\subtable[\label{Tab:PCA4}\sps$N=800$, with PCA.]
{\Tabsize \begin{tabular} {c|@{}c@{}c@{}c@{}c@{}}

                &\ \ $u_1$\ \  & \ \ $u_2$\ \  &\ \  $u_3$\ \  &\ \  $u_4$\ \ \\
   \hline
          $u_1$ &  \ \  $0$    \ \ & \ \ $\underline{0.001}$ \ \  &\ \  $0.258$\ \   &\ \  $0.197$\ \ \\
          $u_2$ &       &\ \ $0$  \ \    &\ \ $\underline{0.000}$\ \    &\ \  $\underline{0.008}$\ \ \\
          $u_3$ &       &       &\ \  $0$    \ \   & \ \ $\underline{0.000}$\ \ \\
          $u_4$ &       &       &       &\ \ $0$   \ \
   \end{tabular}}

     \qquad\hspace{-7mm}
   \subtable[\label{Tab:ICA1}\sps$N=100$, with ICA.]
{\Tabsize \begin{tabular} {c|@{}c@{}c@{}c@{}c@{}}

                &\ \  $u_1$\ \  & \ \ $u_2$\ \  &\ \  $u_3$\ \  & \ \ $u_4$\ \ \\
   \hline
          $u_1$ & \ \  $0$    \ \   &\ \  ${0.080}$\ \   &\ \  $0.098$\ \   &\ \  $0.104$\ \ \\
          $u_2$ &       &\ \ $0$   \ \    &\ \ ${0.095}$ \ \   &\ \  ${0.092}$\ \ \\
          $u_3$ &       &       &\ \   $0$  \ \     &\ \  $0.086$\ \ \\
          $u_4$ &       &       &       & \ \ $0$  \ \
   \end{tabular}}
   \qquad\hspace{-7mm}
   \subtable[\label{Tab:ICA2}\sps$N=200$, with ICA.]
{\Tabsize \begin{tabular} {c|@{}c@{}c@{}c@{}c@{}}

                & \ \ $u_1$ \ \ & \ \ $u_2$ \ \ &\ \  $u_3$ \ \ &\ \  $u_4$\ \ \\
   \hline
          $u_1$ &  \ \ $0$    \ \   &\ \  $0.124$\ \   &\ \  $0.126$\ \   &\ \  ${0.136}$\ \ \\
          $u_2$ &       &\ \ $0$  \ \     &\ \ $0.142$ \ \   & \ \ $0.145$\ \ \\
          $u_3$ &       &       &\ \ $0$    \ \     & \ \ ${0.108}$\ \ \\
          $u_4$ &       &       &       &\ \ $0$   \ \
   \end{tabular}}
   \qquad\hspace{-7mm}
\subtable[\label{Tab:ICA3}\sps$N=400$, with ICA.]
{\Tabsize \begin{tabular} {c|@{}c@{}c@{}c@{}c@{}}

                &\ \  $u_1$\ \  &\ \  $u_2$\ \  &\ \  $u_3$\ \  &\ \  $u_4$\ \ \\
   \hline
          $u_1$ & \ \ $0$    \ \    &\ \  $0.073$\ \   &\ \  $0.222$\ \   & \ \ $0.324$\ \ \\
          $u_2$ &       &\ \ $0$    \ \    &\ \ $0.123$\ \    &\ \  $0.134$\ \ \\
          $u_3$ &       &       & \ \ $0$   \ \     &\ \  $0.155$\ \ \\
          $u_4$ &       &       &       & \ \ $0$   \ \
   \end{tabular}}
   \qquad\hspace{-7mm}
\subtable[\label{Tab:ICA4}\sps$N=800$, with ICA.]
{\Tabsize \begin{tabular} {c|@{}c@{}c@{}c@{}c@{}}

                & \ \ $u_1$\ \  & \ \ $u_2$\ \  &\ \  $u_3$ \ \ & \ \ $u_4$\ \ \\
   \hline
          $u_1$ & \ \ $0$    \ \    &\ \  $0.091$\ \   &\ \  $0.241$ \ \  &\ \  $0.174$\ \ \\
          $u_2$ &       &\ \ $0$   \ \    &\ \ $0.329$\ \    &\ \  $0.353$\ \ \\
          $u_3$ &       &       &\ \ $0$     \ \    &\ \  $0.114$\ \ \\
          $u_4$ &       &       &       &\ \  $0$  \ \
   \end{tabular}}
   \vspace{-5mm}
   \end{table*}

\subsubsection{Decorrelation Performance}
\label{Sec:Decorrelation Performance}
We generated different amounts of samples from a single Dirichlet distribution, where the parameters are chosen to be $\boldsymbol \alpha = [2,5,6,3,7]^{\text{T}}$. The proposed PNT method, which was shown more efficient than the SNT method, was applied to decorrelate the generated samples. With different amounts of data, the DCs between possible pairs of all the transformed variables were evaluated and the corresponding $p$-values are listed in Tab.~\ref{Tab:PNT1},~\ref{Tab:PNT2},~\ref{Tab:PNT3}, and~\ref{Tab:PNT4}, respectively. To make extensive comparison, we also applied the PCA-based decorrelation method and the ICA-based decorrelation method, respectively, to the generated data and summarized the decorrelation performance in Tab.~\ref{Tab:PCA1}-Tab.~\ref{Tab:ICA4}.

When the amount of samples is small (\emph{e.g.}, $N=100$), the generated data cannot reveal neutrality completely (\emph{e.g.}, in Tab.~\ref{Tab:ORG1}, the $p$-value for the DC between $x_1$ and $x_2$ is larger than $0.05$. This indicates that these two variables are independent of each other, which is in conflict with the definition of neutrality.), PNT, PCA, and ICA methods can decorrelate the ``semi''-neutral vector variable into a set of mutually independent scalar variables. As the amount of sample increases, the neutrality of the data becomes clear (\emph{i.e.}, all the $p$-values are smaller than $0.05$ in Tab.~\ref{Tab:ORG2},~\ref{Tab:ORG3}, and~\ref{Tab:ORG4}). It can be observed that both the PNT and the ICA algorithms can yield mutually independent variables for all the cases ($p$-value is larger than $0.05$). In contrast, the PCA algorithm can only lead to partially mutual independence.

In summary, the proposed strategy can nonlinearly transform the highly negatively correlated neutral vector variable into a set of mutually independent scalar variables. Compared with PCA, PNT and ICA show better decorrelation performance for the data with neutral property, with a wide range of amounts of samples. In order to remove the effect of randomness, we ran $50$ rounds of simulations and the mean values are reported in Tab.~\ref{Tab:SingleDecorrelation}. Each round of simulation includes data generation, PNT decorrelation, PCA decorrelation, ICA decorrelation, and DC calculation.

\subsubsection{Discussions}
We compared the computational complexities of SNT, PNT, PCA, and ICA in Sec.~\ref{Sec:Computational Complexity}. The proposed SNT and PNT methods have less computational complexity compared to PCA and ICA. In all of these methods, PNT has the least computational complexity. ICA has the largest computational complexity (usually, $M$ is a number larger than $\log K$). At the meantime, it does not have analytically tractable solution and needs many iterations to converge.

When evaluating these methods with decorrelation performance, we only used PNT to represent the proposed nonlinear transformation strategies. It can be observed that both PNT and ICA have good decorrelatoin performance (in terms of mutual independence measured by DC) for neutral vector variables, with a wide range of data amounts. PCA does not perform well for neutral vector variables when $N$ increases.

In summary, for neutral vector variable, PNT performs better than PCA and ICA, in terms of both decorrelation and computational complexity. {Comparing with PNT and PCA, ICA does not have an analytically tractable solution. Therefore, ICA algorithms typically resort to iterative procedures with either difficulties or high computational load. Hence, we compare only PNT and PCA in the following experiments.

\subsection{Synthesized Data Evaluation}
\subsubsection{Mixture of Dirichlet Distributions}
\begin{table*}[!t]
\vspace{-5mm}
\caption{\label{Tab:MixtureDecorrelation}\footnotesize Evaluation of the decorrelation performance on the data generated from a mixture of Dirichlet distributions with $\pi_1=0.3,\ \pi_2=0.7,$ and $\boldsymbol \alpha_1 = [2,5,6,3,7]^{\text{T}},\ \boldsymbol \alpha_2 = [10,2,8,2,18]^{\text{T}}$. The upper row: $p$-values for the data set with $N=50$ samples. The bottom row: $p$-values for the data set with $N=800$ samples. The $p$-values that are smaller than $0.05$ are marked with underline, indicating that the corresponding two random variables are not independent.}
\centering
\subtable[\label{Tab:MixORG1}\sps Whole data set, original.]
{\Tabsize \begin{tabular}{c|@{}c@{}c@{}c@{}c@{}}

                & \ \ $x_1$\ \  & \ \ $x_2$\ \  & \ \ $x_3$ \ \ & \ \ $x_4$\ \ \\
   \hline
          $x_1$ & \ \ $0$ \ \      & \ \ ${0.107}$\ \  & \ \ $\underline{0.021}$\ \     & \ \ $\underline{0.001}$\ \ \\
          $x_2$ &       &\ \ $0$   \ \    &\ \ ${0.246}$ \ \   & \ \ $\underline{0.019}$\ \ \\
          $x_3$ &       &       & \ \ $0$   \ \     & \ \ ${0.359}$\ \ \\
          $x_4$ &       &       &       &\ \  $0$  \ \
   \end{tabular}}
   \qquad\hspace{-6mm}
\subtable[\label{Tab:MixDecorr1}\sps Whole data set, with PNT.]
 {\Tabsize \begin{tabular}{c|@{}c@{}c@{}c@{}c@{}}

                &\ \  $u_1$\ \  &\ \  $u_2$\ \  &\ \  $u_3$\ \  &\ \  $u_4$\ \ \\
   \hline
          $u_1$ & \ \  $0$   \ \    & \ \ $\underline{0.031}$\ \   & \ \ $\underline{0.029}$ \ \  &\ \  $\underline{0.000}$\ \ \\
          $u_2$ &       &\ \  $0$    \ \  &\ \ ${0.321}$ \ \   &\ \  ${0.109}$\ \ \\
          $u_3$ &       &       &\ \  $0$   \ \     &\ \  ${0.147}$\ \ \\
          $u_4$ &       &       &       & \ \ $0$  \ \
   \end{tabular}}
      \qquad\hspace{-6mm}
   \subtable[\label{Tab:MixComp1-1}\sps Cluster $1$, with PNT.]
 {\Tabsize \begin{tabular} {c|@{}c@{}c@{}c@{}c@{}}

                &\ \  $u_1$\ \  &\ \  $u_2$\ \  &\ \  $u_3$\ \  &\ \  $u_4$\ \ \\
   \hline
          $u_1$ & \ \  $0$   \ \    & \ \ ${{0.471}}$\ \   & \ \ ${{0.610}}$ \ \  &\ \  ${0.480}$\ \ \\
          $u_2$ &       &\ \  $0$    \ \  &\ \ ${{0.463}}$ \ \   &\ \  ${{0.513}}$\ \ \\
          $u_3$ &       &       &\ \  $0$   \ \     &\ \  ${{0.422}}$\ \ \\
          $u_4$ &       &       &       & \ \ $0$  \ \
   \end{tabular}}
         \qquad\hspace{-6mm}
   \subtable[\label{Tab:MixComp1-2}\sps Cluster $2$, with PNT.]
 {\Tabsize \begin{tabular} {c|@{}c@{}c@{}c@{}c@{}}

                &\ \  $u_1$\ \  &\ \  $u_2$\ \  &\ \  $u_3$\ \  &\ \  $u_4$\ \ \\
   \hline
          $u_1$ & \ \  $0$   \ \    & \ \ ${0.468}$\ \   & \ \ ${0.410}$ \ \  &\ \  $0.502$\ \ \\
          $u_2$ &       &\ \  $0$    \ \  &\ \ ${0.614}$ \ \   &\ \  ${0.559}$\ \ \\
          $u_3$ &       &       &\ \  $0$   \ \     &\ \  ${0.534}$\ \ \\
          $u_4$ &       &       &       & \ \ $0$  \ \
   \end{tabular}}
            \qquad\hspace{-6mm}
   \subtable[\label{Tab:MixORG2}\sps Whole data set, original.]
{\Tabsize \begin{tabular}{c|@{}c@{}c@{}c@{}c@{}}

                & \ \ $x_1$\ \  & \ \ $x_2$\ \  & \ \ $x_3$ \ \ & \ \ $x_4$\ \ \\
   \hline
          $x_1$ & \ \ $0$ \ \      & \ \ $\underline{0.000}$\ \  & \ \ $\underline{0.000}$\ \     & \ \ $\underline{0.000}$\ \ \\
          $x_2$ &       &\ \ $0$   \ \    &\ \ $\underline{0.001}$ \ \   & \ \ $\underline{0.000}$\ \ \\
          $x_3$ &       &       & \ \ $0$   \ \     & \ \ $\underline{0.023}$\ \ \\
          $x_4$ &       &       &       &\ \  $0$  \ \
   \end{tabular}}
   \qquad\hspace{-6mm}
\subtable[\label{Tab:MixDecorr2}\sps Whole data set, with PNT.]
 {\Tabsize \begin{tabular}{c|@{}c@{}c@{}c@{}c@{}}

                &\ \  $u_1$\ \  &\ \  $u_2$\ \  &\ \  $u_3$\ \  &\ \  $u_4$\ \ \\
   \hline
          $u_1$ & \ \  $0$   \ \    & \ \ $\underline{0.000}$\ \   & \ \ $\underline{0.000}$ \ \  &\ \  $\underline{0.000}$\ \ \\
          $u_2$ &       &\ \  $0$    \ \  &\ \ $\underline{0.000}$ \ \   &\ \  $\underline{0.000}$\ \ \\
          $u_3$ &       &       &\ \  $0$   \ \     &\ \  $\underline{0.000}$\ \ \\
          $u_4$ &       &       &       & \ \ $0$  \ \
   \end{tabular}}
      \qquad\hspace{-6mm}
   \subtable[\label{Tab:MixComp2-1}\sps Cluster $1$, with PNT.]
 {\Tabsize \begin{tabular} {c|@{}c@{}c@{}c@{}c@{}}

                &\ \  $u_1$\ \  &\ \  $u_2$\ \  &\ \  $u_3$\ \  &\ \  $u_4$\ \ \\
   \hline
          $u_1$ & \ \  $0$   \ \    & \ \ ${{0.529}}$\ \   & \ \ ${{0.484}}$ \ \  &\ \  ${0.429}$\ \ \\
          $u_2$ &       &\ \  $0$    \ \  &\ \ ${{0.511}}$ \ \   &\ \  ${{0.630}}$\ \ \\
          $u_3$ &       &       &\ \  $0$   \ \     &\ \  ${{0.469}}$\ \ \\
          $u_4$ &       &       &       & \ \ $0$  \ \
   \end{tabular}}
         \qquad\hspace{-6mm}
   \subtable[\label{Tab:MixComp2-2}\sps Cluster $2$, with PNT.]
 {\Tabsize \begin{tabular} {c|@{}c@{}c@{}c@{}c@{}}

                &\ \  $u_1$\ \  &\ \  $u_2$\ \  &\ \  $u_3$\ \  &\ \  $u_4$\ \ \\
   \hline
          $u_1$ & \ \  $0$   \ \    & \ \ ${0.483}$\ \   & \ \ ${0.459}$ \ \  &\ \  $0.414$\ \ \\
          $u_2$ &       &\ \  $0$    \ \  &\ \ ${0.531}$ \ \   &\ \  ${0.474}$\ \ \\
          $u_3$ &       &       &\ \  $0$   \ \     &\ \  ${0.517}$\ \ \\
          $u_4$ &       &       &       & \ \ $0$  \ \
   \end{tabular}}
   \vspace{-5mm}
   \end{table*}
In real applications, the data we obtained are usually multimodally distributed. The neutral vector variable is, however, uni-modally distributed by definition. Hence, it is of sufficient interest to study the decorrelation performance of the proposed method on the data sampled from a mixture of Dirichlet distributions. In this section, we generated a set of data from a mixture of Dirichlet distributions to evaluate the decorrelation performance. The chosen model contains two mixture components, which has mixture coefficients as $\pi_1=0.3,\ \pi_2=0.7,$ and component parameters as $\boldsymbol \alpha_1 = [2,5,6,3,7]^{\text{T}},\ \boldsymbol \alpha_2 = [10,2,8,2,18]^{\text{T}}$. Table~\ref{Tab:MixtureDecorrelation} shows the decorrelation performance on the whole data set. The upper row illustrates the decorrelation performance for the data set with $N=50$ samples. As mentioned in the previous section, small amount of data from a single component cannot completely reveal the neutrality. Hence, the data generated from a mixture of Dirichlet distributions may still
have mutual independence between some pairs of dimensions (\emph{e.g.}, in Tab.~\ref{Tab:MixORG1}, the $p$-value for the DC between $x_2$ and $x_3$ is larger than $0.05$, which indicates mutual independence.) In such case, when applying the PNT algorithm to the whole data set, it yields only~\emph{partially} mutual independence (see Tab.~\ref{Tab:MixDecorr1}). For each data cluster, the PNT algorithm works well, as expected (see Tab.~\ref{Tab:MixComp1-1} and~\ref{Tab:MixComp1-2}). With large amount of data ($N=800$), the data generated from each mixture component have strong neutral property so that the whole data set are highly correlated but~\emph{not} neutral (see Tab.~\ref{Tab:MixORG2}). In this case, the PNT algorithm does not work (see Tab.~\ref{Tab:MixDecorr2}). This is because the proposed decorrelation strategy is based on the assumption of neutrality and it may not work for the data that are not neutral. However, if we partition the data into clusters where each cluster contains data vectors that are neutral, the PNT algorithm can perfectly leads to mutual independence between any possible pairs of decorrelated dimensions (see Tab.~\ref{Tab:MixComp2-1} and~\ref{Tab:MixComp2-2}).
\subsubsection{Coding Gain/Removal of Memory Advantage}
\label{Sec:CodingGain}
One advantage of the proposed nonlinear transformation strategy occurs in high rate quantization of vectors. In the application of source coding, the source vectors are usually highly correlated. Hence, it is natural to decorrelate the vector into a set of mutually independent scalars so that the vector quantization (VQ) can be replaced by a set of scalar quantization (SQ) without losing the memory advantage~\cite{Kleijn2010}. This can be quantified by the so-called coding gain measurement~\cite{Gersho1991,Kleijn2010}. For different quantization methods, the coding gain can be measured as (or proportional to) the ratio of quantization distortions, with a given number of bits for quantization.

As shown in~\cite{Gardner1995}, with the high rate assumption, the distortion incurred by quantizing a vector approaches a
simple quadratically weighted error as
\begin{equation}
\eqs
d(\mathbf{x},\widehat{\mathbf{x}})=\left(\mathbf{u}-\widehat{\mathbf{u}}\right)^{\text{T}} \mathcal{J}_{\mathcal{T}}^{\text{T}}(\mathbf{u}) \mathcal{J}_{\mathcal{T}}(\mathbf{u})\left(\mathbf{u}-\widehat{\mathbf{u}}\right),
\end{equation}
where $\mathcal{J}_{\mathcal{T}}$ is the Jacobian matrix of the~\emph{inverse} PNT algorithm
$\mathbf{x}=\mathcal{T}(\mathbf{u})$. The distortion in the $\mathbf{x}$ domain, incurred by quantizing $\mathbf{u}$, can be
approximated as~\cite{Ma2013}
\begin{equation}
\eqs
\begin{split}
\label{DistortionTransformation}
&D_{\mathbf{x}}(\mathbf{u})\cong \sum_{k=1}^{K}\mathbf{E}\left[\mathcal{J}_{\mathcal{T}}^{\text{T}}(\mathbf{u}) \mathcal{J}_{\mathcal{T}}(\mathbf{u})\right]_{k,k}\times D(u_k),
\end{split}
\end{equation}
where $K$ is the dimensionality of $\mathbf{u}$ and $\mathbf{E}[\cdot]$ denotes expectation operation.
In the above  equation, we denote $D(u_k)$ as the distortion incurred by quantization of $u_k$ in the $\mathbf{u}$ domain. By
assuming that $\mathbf{x}$ is Dirichlet distributed with known parameters, we can apply the PNT algorithm to transform
$\mathbf{x}$ to $\mathbf{u}$, and $u_k$ is beta distributed (see~\eqref{PDF4Beta})~\cite{Ma2013}. With the high rate theory and entropy constrained
quantization~\cite{Kleijn2010}, we can derive that, with $R$ bits and probability density function (PDF)-optimized bit allocation strategy~\cite{Ma2010}, the
distortion in the $\mathbf{x}$ domain  incurred by quantizing $\mathbf{u}$ is~\cite{Ma2013}
\begin{equation*}
\eqs
\label{DiffEntropyBeta}
\begin{split}
D_\mathbf{x}(\mathbf{u})&=\frac{K}{12}\times 2^{-\frac{2}{K}\times\left[R - \sum_{k=1}^K h(u_k)\right]}\times \sqrt[K]{\prod_{k=1}^K \mathbf{E}\left[\mathcal{J}_{\mathcal{T}}^{\text{T}}(\mathbf{u})
\mathcal{J}_{\mathcal{T}}(\mathbf{u})\right]_{k,k}}\ ,
\end{split}
\end{equation*}
where $h(u_k)$ is the differential entropy of $u_k$.

On the other hand, if we quantize each element in $\mathbf{x}$ according to its marginal distribution~(this means we replace a vector quantizer by a set of scalar quantizer without decorrelation), the distortion is
\begin{equation}
\eqs
\label{DiffEntropyDir}
D_\mathbf{x}(\mathbf{x})=\frac{K}{12}\times 2^{-\frac{2}{K}\times\left[R - \sum_{k=1}^K h(x_k)\right]}.
\end{equation}
For a $(K+1)$-dimensional Dirichlet distribution with parameter $\boldsymbol \alpha=[\alpha_1,\alpha_2,\ldots,\alpha_{K+1}]^T$, the marginal distribution for the $k$th dimension is
\begin{equation}
\eqs
x_k\sim \mathbf{Beta}(x_k;\alpha_k, \sum_{i=1,i\neq k}^{K+1}\alpha_i).
\end{equation}
Thus we can measure the coding gain as the ratio of two distortions
\begin{equation}
\eqs
\label{CG}
G = \frac{D_\mathbf{x}(\mathbf{x})}{D_\mathbf{x}(\mathbf{u})}=\frac{2^{\frac{2}{K}\sum_{k=1}^K \left[h(x_k)-h(u_k)\right]}}{\sqrt[K]{\prod_{k=1}^K \mathbf{E}\left[\mathcal{J}_{\mathcal{T}}^{\text{T}}(\mathbf{u})
\mathcal{J}_{\mathcal{T}}(\mathbf{u})\right]_{k,k}}}.
\end{equation}
In the above equation, the ratio $G>1$ indicates less distortion can be achieved by the proposed nonlinear transformation. The larger this ratio is, the more benefit we obtain from the transformation. In order to evaluate the coding gain $G$ extensively, we evaluated the coding gain with different
$\boldsymbol \alpha$ and different dimensionalities. To give an example, the inverse nonlinear transformation and the elements in $\mathcal{J}_{\mathcal{T}}(\mathbf{u})$ with $K=4$ are listed in Tab.~\ref{Tab:JacobianK4}. The expectation term in the denominator of~\eqref{CG} can be calculated in a closed-form expression with the fact that $u_i$ is beta distributed and the parameters can be calculated from the original Dirichlet parameters (see~\eqref{PDF4Beta} for more details).

The coding gains with $K=4,5,6$ are plotted in Fig.~\ref{Fig: CodingGain}. For each $K$, we randomly generated the elements in $\boldsymbol \alpha$ from $[10,50]$. In total $100$ rounds of simulations were conducted for each $K$. It can be observed that the proposed nonlinear transformation yield a coding gain greater than $1$ for different dimensions. This is because the memory advantage of VQ over SQ has been removed.
\begin{table*}[!t]
\vspace{-5mm} \caption{\label{Tab:JacobianK4}\small The inverse PNT algorithm $\mathbf{x}=\mathcal{T}(\mathbf{u})$ and the Jacobian matrix $\mathcal{J}_{\mathcal{T}}(\mathbf{u})$ for $(K=4)$.}
\centering \vspace{0mm}
\Tabsize
\begin{tabular}{|c|c|}
\hline
$\mathbf{x}=\mathcal{T}(\mathbf{u}):\begin{array}{c}
x_1=u_1u_3u_4\\
x_2=(1-u_1)u_3u_4\\
x_3=u_2(1-u_3)u_4\\
x_4=(1-u_2)(1-u_3)u_4
\end{array}$
&
$\mathcal{J}_{\mathcal{T}}(\mathbf{u})=\left[\begin{array}{cccc}
u_3u_4&0&u_1u_4&u_1u_3\\
-u_3u_4&0&(1-u_1)u_4&(1-u_1)u_3\\
0&(1-u_3)u_4&-u_2u_4&u_2(1-u_3)\\
0&-(1-u_3)u_4&-(1-u_2)u_4&(1-u_2)(1-u_3)
\end{array}\right]$\\
\hline
\end{tabular}
\vspace{0mm}
\end{table*}
\subsubsection{Discussion}
\label{Sec: Disc}

\begin{figure}[!t]
\vspace{-0mm}
\psfrag{a}[c][c]{ \sps$K=4$}

\psfrag{b}[c][c]{ \sps$K=5$}

\psfrag{c}[c][c]{ \sps$K=6$}

\vspace{0mm}
          \centering
          \includegraphics[width=.3\textwidth]{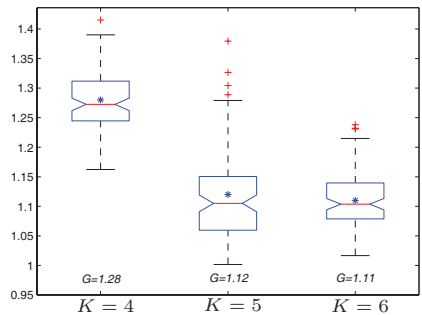}
          \vspace{-2mm}
          \caption{\label{Fig: CodingGain}\footnotesize Coding gains for different $K$ shown as box plot. The central red mark is the median, the blue star mark is the mean, the edges of the box are the $25^{th}$ and $75^{th}$ percentiles. The outliers are marked with red crosses. The mean values are listed at bottom.}
          \vspace{-4mm}
\end{figure}
The synthesized data experiments above demonstrated the superior performance of the proposed nonlinear transformation strategy for neutral data. The data generated from a mixture of Dirichlet distributions are multimodally distributed so that they are not neutral. In this case, we can partition the data into different clusters. By assuming that the data assigned to each cluster were generated from a single Dirichlet distribution, the proposed method can be applied to these data and results in promising decorrelation performance. Decorrelation of highly negatively correlated vector plays an important role in many applications. In the next section, we will apply this idea to real data applications.

\subsection{Real Data Evaluation}
Decorrelation of a highly correlated vector variable into a set of mutually independent variables leads to many advantages in
real applications~\cite{Bishop2006,Ma2013,Kleijn2010,Torre2012,Han2014}. In this section, we evaluate the decorrelation performance of the proposed strategy for real life data that fit the definition of neutral vector (nonnegative and $l_1$ norm equals one).
To this end, we assume such ``neutral-like''\footnotemark\footnotetext{Hereby, we name the vector 1) contains nonnegative elements and 2) has unit/constant $l_1$-norm as ``neutral-like'' data.} data have neutral property and apply the PNT algorithm to nonlinearly transform them. The performance improvement in practical applications is also presented.
\subsubsection{\textbf{Vector Quantization of Line Spectral Frequency Parameters}}
Quantization of the LSF parameters of the linear predictive coding (LPC) model is an essential part of speech transmission~\cite{Ma2013,Lee2011,Wang2015}. The LSF parameters are usually $10$-dimensional for narrow band speech and $16$-dimensional for wide band speech. Hence, vector quantization (VQ) is required. Generally speaking, VQ has memory, shape, and space-filling advantages over scalar quantization (SQ)~\cite{Kleijn2010,Lee2011}. However, it is~\emph{impractical} to design a full vector quantizer because 1) the size of codebook increases exponentially with the dimension of data, which leads to high storage complexity; 2) the effort of training a codebook and searching for an index in the codebook is also exponentially increased with the data's dimension, which is computationally costly. Especially, when the dimension is high,~\emph{e.g.},~$>10$, the above VQ is not feasible. In practical VQ implementation, the frequently used method is to decorrelate the LSF parameters into a set of mutually independent scalars so that the memory advantage of VQ over SQ can be removed~\cite{Kleijn2010,Gersho1991,Lee2011}. Then, a set of SQs will be employed to replace the VQ.

In the design of PDF-optimized VQ, the Gaussian distribution and the corresponding Gaussian mixture model (GMM) have been intensively applied to model the distribution of the LSF parameters~\cite{Subramaniam2003,Chatterjee2008,Ramirez2013}. However, since the LSF parameters are in the interval $(0,\pi)$ and are strictly ordered, it is~\emph{not} Gaussian distributed. For the purpose of more efficient modeling, the LSF parameters can be converted to the so-called $\Delta$LSF parameters~\cite{Ma2010,Ma2013}. The $\Delta$LSF parameters are nonnegative and the summation equals $1$\footnotemark\footnotetext{Strictly speaking, the summation of the $\Delta$LSF parameters equals $\pi$, which can be scaled so that the summation equals $1$. The scaled $\Delta$LSF parameters represent the proportions of the $\Delta$LSF on the unit circle~\cite{Ma2013}.}. As the $\Delta$LSF parameters fit the the definition, we suppose that they follow Dirichlet distributions and apply a Dirichlet mixture model (DMM) to describe the underlying distribution of the data. As data generated from a Dirichlet distribution have neutral property, the proposed nonlinear strategy is applied to decorrelate the $\Delta$LSF parameters. A practical VQ is carried out based on the neutrality.

\begin{figure}[!t]
\vspace{-0mm}
\psfrag{A}[c][c]{ \sps Independence coefficient}
\psfrag{c}[c][c]{ \sps $C_1$}
\psfrag{d}[c][c]{ \sps $C_2$}
\psfrag{e}[c][c]{ \sps $C_3$}
\psfrag{f}[c][c]{ \sps $C_4$}
\psfrag{g}[c][c]{ \sps $C_5$}
\psfrag{h}[c][c]{ \sps $C_6$}
\psfrag{i}[c][c]{ \sps $C_7$}
\psfrag{j}[c][c]{ \sps $C_8$}
\vspace{0mm}
          \centering
          \includegraphics[width=.3\textwidth]{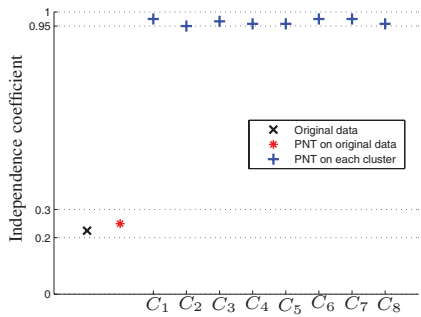}
          \vspace{-2mm}
          \caption{\label{Fig:PNTonRealData800}\footnotesize Independence coefficients of different data set. $C_i$ denotes the $i^{th}$ cluster obtained by the EM algorithm. The amount of samples is $N=800$ and the number of mixture components (clusters) is $8$.}
          \vspace{-5mm}
\end{figure}
\begin{itemize}
\item{\emph{Evaluation of Independence}}

 The $\Delta$LSF parameters are $16$-dimensional\footnotemark\footnotetext{We show only the results for wide band data here. Similar performance can also be obtained for narrow band data.} for wide band speech data. It is space consuming to list a $16\times 16$ mutual independence $p$-value table. Thus, we calculated~\emph{independence coefficient}~(IC), which is defined as the proportion of the number of mutually independent pairs to the number of all the possible pairs\footnotemark\footnotetext{For a $K\times K$ matrix, the number of all the possible pair is $\frac{K(K-1)}{2}$, without consideration of self pairs.} to measure the decorrelation performance. The higher this proportion is, the better the decorrelation performance is\footnotemark\footnotetext{The largest ratio is $1$, which means all the possible pairs are mutually independent.}.

 As described in Sec.~\ref{Sec: Disc}, we firstly applied the PNT algorithm to the $\Delta$LSF parameters. As shown in Fig.~\ref{Fig:PNTonRealData800}, the IC of PNT for the original data is small, which means the decorrelation performance of PNT is not significant. This is due to the fact that the $\Delta$LSF parameters are multimodally distributed. We applied the EM algorithm~\cite{Ma2013} to partition the $\Delta$LSF parameters into different clusters. With the assumption that the data in each cluster are Dirichlet distributed (hence, they are neutral vectors), we applied the PNT algorithm to the data in each cluster, respectively. The ICs of PNT for each cluster are also plotted in Fig.~\ref{Fig:PNTonRealData800}. It is clearly shown that most of the pairs (more than $95\%$) are mutually independent. Hence, the mutual correlation for each cluster has been significantly removed by PNT.

\item{\emph{Improvement in VQ}}

Motivated by the coding gain advantage in Sec.~\ref{Sec:CodingGain}, we designed and implemented a DMM-based VQ based on the neutral properties. The LSF parameters were partitioned into $I$\footnotemark\footnotetext{Usually, $I$ equals a power of $2$.} clusters with a DMM which contains $I$ mixture components~\cite{Ma2010}. With the above introduced procedure, the PNT algorithm is applied to realize the decorrelation for each cluster and a set of mutually independent scalar elements are obtained. As the memory advantage of VQ over SQ is removed by explicitly using the neutrality, we carried out a PDF-optimized VQ for the LSF parameters. The benefits are two fold:
\begin{enumerate}
\item \emph{Saving of the storage, training and searching costs.} With average bit rate (in per vector sense) $R$, there are $\log_2 M$ bits spent on indexing the mixture component and $R_q=R-\log_2 M$ bits spent on VQ. Hence, by assuming all the components are identical to each other, a codebook with $2^{{R_q}}$ codewords is required for each mixture component. In the SQ case, the bit for each cluster (\emph{i.e.}, mixture component) will be further placed on each dimension based on its differential entropy. On average, $\frac{R_q}{16}$ is assigned to each dimension and only $16\times 2^{\frac{R_q}{16}}$ is needed for each component. Usually, $R$ is a number about $40\sim 50$. Hence, the required number of codewords is significantly reduced and the storage cost is saved. The well-known Lloyd algorithm~\cite{Lloyd1982,Kim2011} and the Linde-Buzo-Gray (LGB) algorithm~\cite{Linde1980,Koren2005} are usually utilized for obtaining the codebook. In the case of VQ, the training is carried out in a $16$-dimensional space. Meanwhile, the training is executed in one-dimensional space for SQ. Obviously, training a codebook in $16$-dimensional space is more computationally costly than that in one-dimensional space, and, therefore, the training cost is saved. For the same reasoning, the searching cost is also significantly reduced when replacing VQ by SQ.
\begin{figure}[!t]
\vspace{-0mm}
          \centering
          \subfigure[\scriptsize DMM-based VQ.]{\includegraphics[width=.4\textwidth]{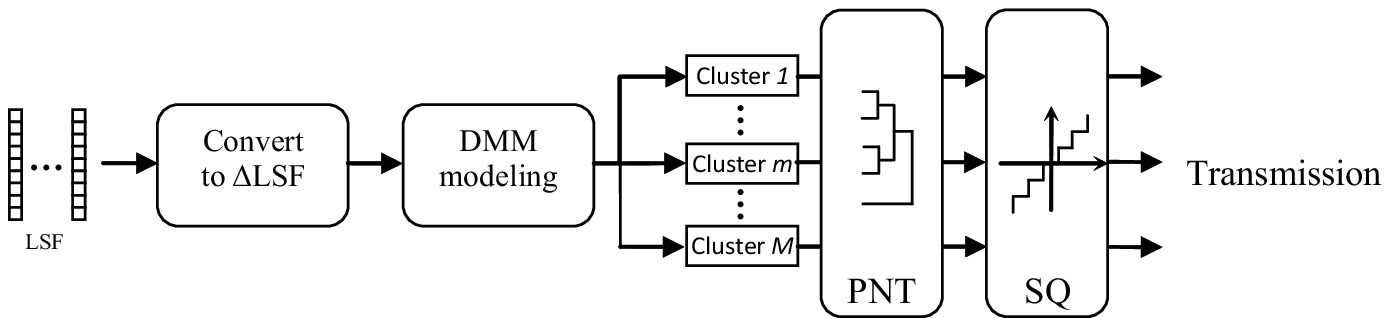}}
          \subfigure[\scriptsize GMM-based VQ.]{\includegraphics[width=.4\textwidth]{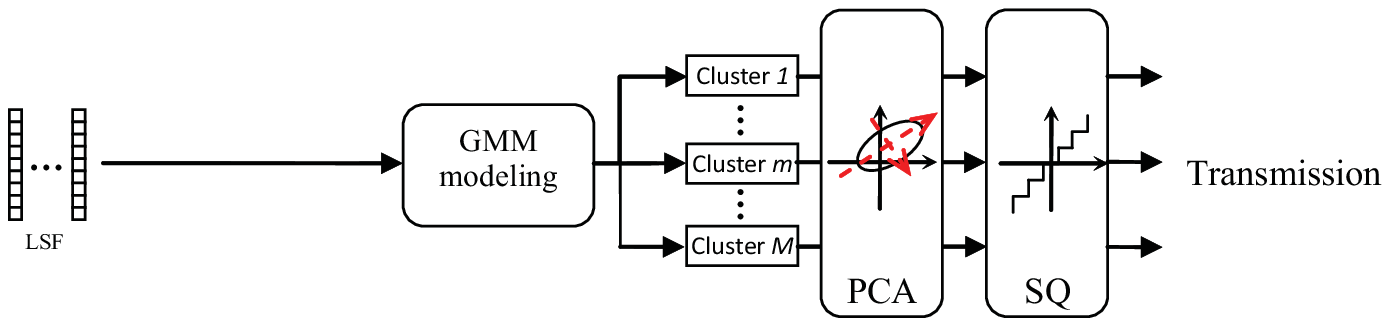}}
          \vspace{-2mm}
          \caption{\label{Fig:DMMVQ}\footnotesize Flow chart of DMM-based VQ and GMM-based VQ.}
          \vspace{-6mm}
\end{figure}
\item \emph{Saving of Bit rates.} The ultimate goal of PDF-optimized VQ is to spend as less bits as possible while satisfying the quantization distortion requirement. A practical VQ for the LSF parameters, which is based on the DMM modeling and the proposed nonlinear transformation strategy, was introduced in~\cite{Ma2013}. With the transparent coding\footnotemark\footnotetext{Transparent coding criterion: 1) $1$ dB LSD on average,
2) less than $2$\% outliers in $2\sim4$ dB range, and 3) no outlier larger than $4$ dB.} criterion, we evaluated the log spectral distortion (LSD) obtained from the DMM-based VQ and compared it with the state-of-the-art GMM-based VQ~\cite{Chatterjee2008a}. The GMM-based VQ partitioned the LSF parameters into $I$ clusters with the EM algorithm for GMM. Next, the LSF parameters are decorrelated with PCA. Finally, a PDF-optimized GMM-based VQ is carried as well. Fig.~\ref{Fig:DMMVQ} shows the designs for the DMM-based VQ and the GMM-based VQ. The VQ performance comparisons are summarized in Tab.~\ref{Tab:VQComparison}. It is clearly demonstrated that the DMM-based VQ improves the performance by about $3$ bits/vector. This is due to the fact that the proposed nonlinear transformation strategy removes the memory advantage and makes the implementation of practical VQ feasible. More details can be found in~\cite{Ma2013}.
\end{enumerate}

\subsubsection{\textbf{EEG Signal Classification}}

\begin{table}[!t]
\vspace{-5mm} \caption{\label{Tab:VQComparison}\footnotesize Comparisons of VQ performance.The number of mixture components is $M=256$. $706k$ LSF vectors were used for training and $258k$ were used for evaluation. The speech data are from the TIMIT database~\cite{TIMIT1990}.}
\centering \vspace{0mm}
\Tabsize
\begin{tabular}{|c|c|c|cc|}
\hline
\multirow{2}{*}{VQ Type}&\multirow{2}{*}{bits/vec.}&\multirow{2}{*}{LSD (dB)}&\multicolumn{2}{|c|}{LSD outliers (in \%)}\\
             &             &                         &$2-4$ dB  &$>4$ dB\\
             \hline
             \hline
\multirow{2}{*}{DMM-based VQ}& $44$ & $1.039$ & $1.200$ & $0.000$\\
                             & $45$ & $0.997$ & $0.830$ & $0.000$\\
             \hline
\multirow{2}{*}{GMM-based VQ}& $47$ & $1.029$ & $0.776$ & $0.005$\\
                             & $48$ & $0.971$ & $0.920$ & $0.003$\\
\hline
\end{tabular}
\vspace{-5mm}
\end{table}
For persons who suffer from neuromuscular diseases, brain-computer interface (BCI) connects them with computers by recording and analyzing the brain signals.
As non-invasively acquired signal, the Electroencephalogram (EEG) signal is the most studied and applied one in the design of a BCI system~\cite{Lotte2007,Cecotti2011}. For the EEG signal obtained from one channel, various types of features have been extracted from the signal for
the purpose of classification. The marginal discrete wavelet transform (mDWT) vector, among others, is a typical feature that is widely adopted~\cite{Subasi2007,Ma2012,Xu2015}.
The elements in a DWT vector reveal features related to the transient nature of the signal. The marginalization operation, which yields the mDWT vector, makes the DWT vector insensitive to time alignment~\cite{Subasi2007}. The data set used in this paper is from the BCI competition III~\cite{BCI}. During one EEG signal trial recording, a subject had to perform imagined
movements of either the left small finger or the tongue. The data set contains $278$ trials for training and $100$ trials for test. The trials in the training and test sets are evenly distributed and labeled, respectively. For each trial, $64$ channel data of length $3000$ samples were provided. The mDWT vector contains nonnegative elements and has unit $l_1$-norm. Hence, we applied the nonlinear transformation method to decorrelate the mDWT vector for the purposed of classification accuracy improvement.

In our previous work~\cite{Ma2016}, we have successfully applied the proposed PNT method in EEG signal classification. The so-called multivariate Beta distribution (mvBeta)-based classifier was introduced based on the feature selection strategy in the transformed feature domain and has been applied to classify the EEG signals. In this paper, we will make thorough study to show that the obtained gain in classification accuracy is indeed from the application of the PNT method to the mDWT vectors.

\begin{itemize}
\item \emph{Channel Selection}

Not all the channels are closely relevant to the classification task. Before conducting the classification task, it is of importance to select more relevant channels so that the classification accuracy can be improved. The Fisher ratio (FR) and the generalization error estimation (GEE)~\cite{Ma2016,Lal2004} were applied to select channels. The channels are ranked according to their FRs and GEEs, respectively. In the classification stage, we exploit the mDWT vectors from the top $m$ channels.

\item \emph{Feature Selection}

Feature selection is an important problem in EEG signal classification~\cite{Ma2016,Subasi2007,Suk2013}. For each selected channel, the dimension of the mDWT vector is $5$ (the degrees of freedom is $4$). We applied the PNT algorithm to decorrelate the mDWT vectors from the training set. A set of $4$-dimensional vectors, each of which contains mutually independent elements were obtained. We sorted the $4$ dimensions according to their variances in descending order. The mDWT vectors from the test set were also decorrelated via PNT. The dimension reordering was carried out based on the variance order from the training set. According to the reordered dimensions, we selected the relevant $D$ dimensions for classification.
\end{itemize}

\item \emph{Performance Improvement}

For binary classification task, the support vector machine (SVM) is a classic and the widely applied classifier~\cite{Cortes1995,Bishop2006,Huang2014,Nan2015}. We evaluated the above introduced feature selection strategy by comparing the classification accuracies. For each channel selection method, an SVM with radial basis function (RBF) kernel was trained as the benchmark, respectively. With LIBSVM toolbox~\cite{Chang2011}, we adjusted the parameters in the RBF-SVM so that the cross validation of training accuracy is the highest.~\emph{All} mDWT vectors from the training set were used for the parameter adjustment. To make fair comparisons, we also applied PCA to decorrelate the mDWT vectors. The mDWT vectors in the test set were transformed with the eigenvectors obtained from the training set. The relevant dimensions were also selected according to the variances.
The classification results were obtained with the top $m$ channels (ranked via FR or GEE). For each channel, the most relevant $D$ features (ranked via variance) were selected. In total, we obtained $(m\times D)$-dimensional feature vector to train the RBF-SVM. It can be observed that the RBF-SVM+PNT yields the highest recognition accuracies, for FR case and GEE case, respectively.
Figure~\ref{Fig:SVM} illustrates the classification results with top $m$ channels when $D=2$. The highest classification rates are both obtained with $D=2$, which indicates that feature selection via variance indeed benefits the classification. The RBF-SVM+PNT yields the highest recognition accuracy for FR case ($75\%$ with with $D=2$ and $m=19,20$) and GEE case ($77\%$ with $D=2$ and $m=4$), respectively.
\end{itemize}

\subsubsection{Discussion}
The LSF parameters in the LPC model and the mDWT parameters in the EEG signal contain nonnegative elements and have unit/constant $l_1$-norm, respectively. Although it is difficult (or even not feasible) to prove the neutrality for such neutral-like data, we can still exploit the neutrality to apply the PNT-based nonlinear transformation strategy for the purpose of decorrelation and improve practical performance. Compared with the PCA-based linear transformation strategy, the PNT-based nonlinear transformation showed advantages in both applications.

\begin{figure}[!t]
\vspace{-0mm}
\psfrag{a}[c][c]{ \tiny Top $m$ channels}
\psfrag{b}[c][c]{ \tiny Classification accuracy (in $\%$)}
          \centering
\begin{tabular}{@{}c@{}}

 \subfigure[\label{SVMD2}\scriptsize Channel selection with FR and $D=2$.]{
          \includegraphics[width=.35\textwidth]{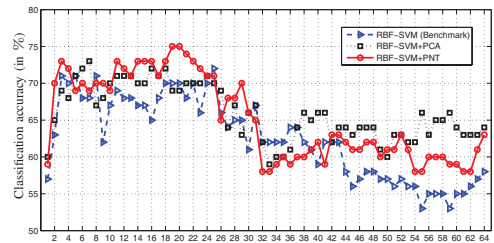}} \\          \subfigure[\label{SVMD2GEE}\scriptsize Channel selection with GEE and $D=2$.]{
          \includegraphics[width=.35\textwidth]{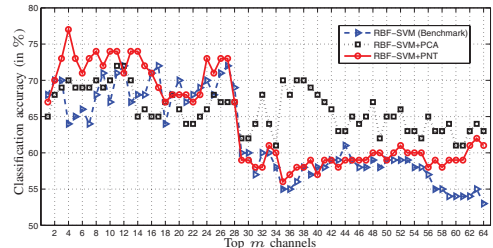}}

\end{tabular}
\vspace{0mm}
\caption{\label{Fig:SVM}\footnotesize Classification accuracy comparisons of RBF-SVM (benchmark, no transformation), RBF-SVM+PCA, and RBF-SVM+PNT with $D=2$. The results have been presented in~\cite{Ma2016}.}
\vspace{-5mm}
\end{figure}

\vspace{0mm}
\section{Conclusions}
\label{Sec:Conclusion}
Nonlinear transformations for neutral vector variable were proposed and studied in this paper. By explicitly utilizing the neutrality of neutral vector variables, we introduced the serial nonlinear transformation and parallel nonlinear transformation methods to decorrelate a neutral vector variable into a set of mutually independent element variables. The mutual independence was theoretically proved. The computational costs of the proposed decorrelation methods were analyzed and compared with the PCA-based and ICA-based approaches. It has been shown that the computational costs of the proposed methods are the smallest.

As a typical case, the vector variable following the Dirichlet distribution is a completely neutral vector. The transformed element variables are all beta distributed. With the distance correlation metric, the decorrelation performance of the proposed nonlinear transformation was demonstrated to be superior to those of PCA and ICA with both synthesized and real life data. Moreover, we applied the proposed nonlinear transformation in two applications,~\emph{i.e.}, quantization of line spectral frequency parameters in the speech linear predictive model and EEG signal classification. Extensive experimental results showed that, when carrying out decorrelation and feature selection for neutral-like data, the proposed parallel nonlinear transformation (PNT)-based nonlinear transformation can achieve better practical performance and is preferable to the conventionally applied PCA-based linear transformation.

\vspace{-3mm}
\section*{Acknowledgement}
This work is partly supported by the National Nature Science Foundation of China (NSFC) grant No. $61402047$, $61511130081$, and $61273217$, Beijing Natural Science Foundation (BNSF) grant No.~$4162044$.

\vspace{-3mm}
\footnotesize\bibliographystyle{IEEEtran}

\end{document}